\PassOptionsToPackage{table,dvipsnames}{xcolor}
\documentclass[]{fairmeta}
\usepackage{amsmath,amsfonts,bm}

\def\eqref#1{equation~\ref{#1}}
\def\1{\bm{1}}

\DeclareMathAlphabet{\mathsfit}{\encodingdefault}{\sfdefault}{m}{sl}
\SetMathAlphabet{\mathsfit}{bold}{\encodingdefault}{\sfdefault}{bx}{n}

\def\gD{{\mathcal{D}}}

\def\gJ{{\mathcal{J}}}

\def\gL{{\mathcal{L}}}
\def\gM{{\mathcal{M}}}

\newcommand{\E}{\mathbb{E}}

\newcommand{\KL}{{\mathrm{KL}}}

\usepackage[utf8]{inputenc}
\usepackage{amssymb}
\usepackage{amsthm}
\usepackage{bbm}
\usepackage{enumitem}
\usepackage{wrapfig}
\usepackage{mathtools}
\usepackage{tabularx}
\usepackage{makecell}
\usepackage[tight]{minitoc}
\usepackage{nicefrac}
\usepackage{fontawesome5}


\newcommand{\mypar}[1]{\noindent \textbf{#1}}

\theoremstyle{definition}
\newtheorem*{remark}{Remark}

\def\policy{\pi_{\theta}}
\def\reference{\pi_{\mathrm{ref}}}
\def\teacher{\pi_{\mathrm{T}}}
\def\teacherbase{\pi_{\mathrm{T}}^{\mathrm{base}}}

\crefformat{section}{\S#2#1#3}
\crefformat{subsection}{\S#2#1#3}
\crefformat{subsubsection}{\S#2#1#3}
\crefformat{paragraph}{\P#2#1#3}
\crefformat{subparagraph}{\P#2#1#3}
\crefmultiformat{section}{\S#2#1#3}{ and~\S#2#1#3}{, \S#2#1#3}{, and~\S#2#1#3}
\crefmultiformat{subsection}{\S#2#1#3}{ and~\S#2#1#3}{, \S#2#1#3}{, and~\S#2#1#3}
\crefmultiformat{subsubsection}{\S#2#1#3}{ and~\S#2#1#3}{, \S#2#1#3}{, and~\S#2#1#3}
\crefmultiformat{paragraph}{\P\P#2#1#3}{ and~#2#1#3}{, #2#1#3}{, and~#2#1#3}
\crefmultiformat{subparagraph}{\P\P#2#1#3}{ and~#2#1#3}{, #2#1#3}{, and~#2#1#3}
\crefrangeformat{section}{\mbox{\S\S#3#1#4--#5#2#6}}
\crefrangeformat{subsection}{\mbox{\S\S#3#1#4--#5#2#6}}
\crefrangeformat{subsubsection}{\mbox{\S\S#3#1#4--#5#2#6}}
\crefrangeformat{paragraph}{\mbox{\P\P#3#1#4--#5#2#6}}
\crefrangeformat{subparagraph}{\mbox{\P\P#3#1#4--#5#2#6}}
\crefname{part}{Part}{Parts}
\Crefname{part}{Part}{Parts}
\crefname{chapter}{Ch.}{Ch.}
\Crefname{chapter}{Ch.}{Ch.}
\crefname{footnote}{Fn.}{Fn.}
\Crefname{footnote}{Fn.}{Fn.}
\crefname{figure}{Figure}{Figures}
\crefname{table}{Table}{Tables}
\crefname{subfigure}{Figure}{Figures}
\Crefname{subfigure}{Figure}{Figures}
\crefname{equation}{Equation}{Equations}
\Crefname{equation}{Equation}{Equations}
\crefname{appsec}{Appendix}{Appendices}
\Crefname{appsec}{Appendix}{Appendices}
\crefname{algocf}{Algorithm}{Algorithms}
\Crefname{algocf}{Algorithm}{Algorithms}

\definecolor{lblue}{HTML}{A6CEE3}
\definecolor{lgreen}{HTML}{D1FFBD}
\definecolor{lred}{HTML}{FAA0A0}
\definecolor{lorange}{HTML}{FDBF6F}
\definecolor{mblue}{HTML}{80B1D3}
\definecolor{mgreen}{HTML}{B3DE69}
\definecolor{mred}{HTML}{FB8072}
\definecolor{morange}{HTML}{FDB462}
\definecolor{blue}{HTML}{1F78B4}
\definecolor{green}{HTML}{33A02C}
\definecolor{red}{HTML}{E31A1C}
\definecolor{orange}{HTML}{FF7F00}
\definecolor{dblue}{HTML}{08519C}
\definecolor{dgreen}{HTML}{006D2C}
\definecolor{dorange}{HTML}{EC7014}

\definecolor{darkblue}{rgb}{0,0,0.5}
\definecolor{darkpink}{rgb}{0.8, 0.125, 0.6}
\setlabdisplayname{OmniAI Group of ZJU ACES Lab}
\setuniversityname{}

\title{Scaling properties of same-family on-policy distillation}

\author[1,2\dagger]{Yuntai Bao}
\author[1]{Qinfeng Li}
\author[2]{Guoqing Jiang}
\author[2]{Liwei Chen}
\author[2]{Zhiheng Qin}
\author[2]{Xuanping Li}
\author[1]{Wenqi Zhang}
\author[1\ddagger]{Xuhong Zhang}
\affiliation[1]{Zhejiang University}
\affiliation[2]{Kuaishou Technology}
\contribution[\dagger]{Work done during internship at Kuaishou Technology}
\contribution[\ddagger]{Corresponding author}
\metadata[\faEnvelope\ Email]{\email{yuntaibao@zju.edu.cn}; \email{zhangxuhong@zju.edu.cn}}
\date{September 2026}

\hypersetup{
  pdftitle={Scaling properties of same-family on-policy distillation},
  pdfauthor={Yuntai Bao, Qinfeng Li, Guoqing Jiang, Liwei Chen, Zhiheng Qin, Xuanping Li, Wenqi Zhang, Xuhong Zhang},
  pdfsubject={Scaling properties of on-policy distillation for capability transfer across model scales},
  pdfkeywords={on-policy distillation, scaling laws, weak-to-strong generalization, reinforcement learning, large language models}
}

\abstract{%
\textit{Reinforcement learning (RL)} can induce substantial reasoning capabilities in large language models (LLMs), but how much of this capability transfers across model scales, and how quickly, remains unclear.
We study the scaling properties of \textit{on-policy distillation (OPD)} across \textit{weak-to-strong}, \textit{same-base}, and \textit{strong-to-weak} teacher--student setups.
We find that early OPD training dynamics uniformly exhibit a regular \textit{useful-transfer} regime, in which held-out accuracy (the \textit{gold score}, $G$) rises approximately linearly in $d\!=\!\smash{\sqrt{\KL(\policy \Vert \reference)}}$, the square root of token-level reverse KL divergence from the student initialization.
In every observed weak-to-strong pair, the student's peak gold score exceeds its teacher's own, so a compact RL expert can transfer capability to a much larger student via OPD.
To estimate OPD outcomes, we fit \textit{power laws} for how $G_{\mathrm{peak}}$ and the slope of the useful-transfer regime scale with student and teacher parameter counts and with teacher gold score.
These laws show that peak gold score improves with teacher scale only up to roughly the student's scale, and that at a matched gold score smaller teachers transfer better, so a teacher's score alone does not define its supervision value.
We also study the scaling effects of two OPD variants, bootstrapping weak-to-strong OPD, and the degree of on-policy supervision.
}

\begin{document}

\doparttoc
\faketableofcontents

\maketitle
\thispagestyle{firstheader}

%%%%%%%%%%%%%%%%%%%%%%%%%%%%%%%%%%%%%%%%%%%%%%%%%%%%%%%%%%%%%%%%%%%%%%%%%%%%%%%
%%%%%%%%%%%%%%%%%%%%%%%%%%%%%%%%%%%%%%%%%%%%%%%%%%%%%%%%%%%%%%%%%%%%%%%%%%%%%%%
\section{Introduction}
\label{sec:introduction}

Modern reasoning models often acquire task expertise through \textit{reinforcement learning (RL)}~\citep{shao2024deepseekmath,olmo2025olmo,yang2025qwen3}.
\textit{On-policy distillation (OPD)} is increasingly used in LLM post-training pipelines to transfer such expertise between models: a teacher policy provides token-level supervision on rollouts sampled from the student, without forcing the student to imitate the teacher's own trajectories~\citep{agarwal2024onpolicy,gu2024minillm,lu2025onpolicydistillation}.
Recent work has produced a growing family of OPD variants that reshape token rewards and distillation objectives~\citep{song2026survey,yang2026learning,ko2026relaxed}, and empirical analyses examine OPD training dynamics along axes such as rollout length and token-overlap ratio~\citep{fu2026revisiting,li2026rethinking}.
However, the scaling properties of OPD, especially how outcomes depend on teacher and student scale, remain poorly understood.
If these dependencies were captured in law-like form, the outcome of an OPD run could be predicted rather than discovered after training.
Therefore, \textbf{the central question} we ask is:
\textit{%
can we estimate the performance of the outcome student policy from teacher and student scale, before performing OPD?}

Studies of reward model overoptimization offer an empirical roadmap: in RL with \textit{proximal policy optimization (PPO)} against a learned reward model, gold reward dynamics can be described as functions of the KL divergence from the initial policy, with coefficients predictable from reward-model scale~\citep{gao2023scaling}.
A similar phenomenon arises in direct alignment algorithms such as DPO~\citep{rafailov2024scaling}.
OPD also optimizes the student against a proxy, the token-level implicit reward induced by a fixed teacher.
We adopt the same approach: we characterize held-out accuracy (the \textit{gold score} $G$, by analogy with the gold reward above) along the KL-indexed training progress $d\coloneqq\smash{\sqrt{\KL(\policy \Vert \reference)}}$ from the initial student.
We then try to predict the fitted coefficients from student and teacher scale.
We find that OPD dynamics share an initial regular \textit{useful-transfer regime} where gold score rises approximately linearly with $d$.
Subsequent dynamics are noisy and separate into attenuated improvement, saturation, and regression, unlike the regular overoptimization regime of PPO and direct alignment.

\begin{figure}[t]
\centering
\includegraphics[width=.98\linewidth]{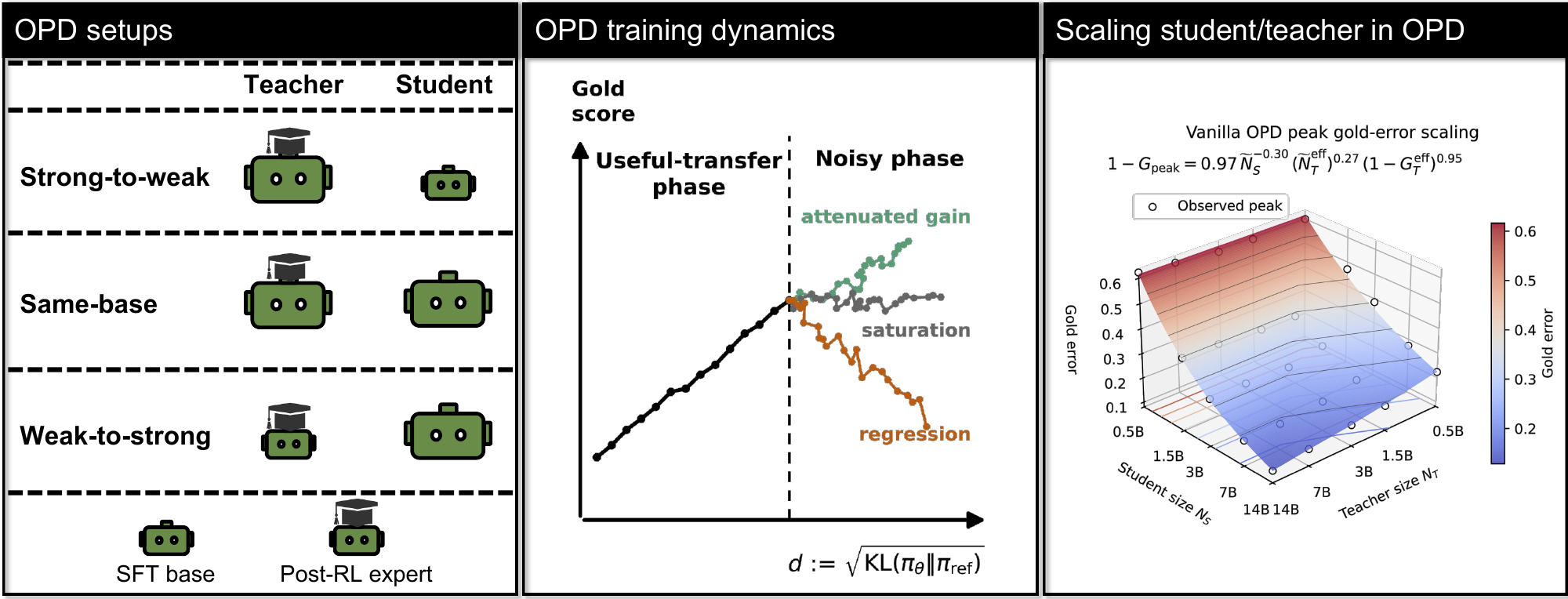}
\caption{OPD transfers expertise from an RL-trained teacher to a student policy.
\textbf{Left:} the three teacher--student setups, with post-RL experts as teachers and SFT bases as students.
\textbf{Center:} gold score rises linearly with KL-indexed training progress $d$ during the useful-transfer phase, after which dynamics turn noisy.
\textbf{Right:} the fitted joint power law predicts peak gold error from teacher and student scale.}
\label{fig:overview}
\end{figure}

OPD applies across three teacher--student configurations: \textit{strong-to-weak}, \textit{same-base}, and \textit{weak-to-strong}.
The weak-to-strong direction is especially attractive because it amortizes task-specific post-training across a model family: instead of repeating RL at every scale, a weak expert serves as a cheap proxy for task expertise, while the strong student contributes knowledge and reasoning strategies its teacher lacks.
OPD strengthens prior weak-to-strong supervision~\citep{burns2024weak,yuan2026incentivizing} by querying the teacher on fresh rollouts from the evolving student rather than relying on teacher demonstrations.
Recent weak-to-strong OPD methods show that policy contrasts can improve transfer from weak RL experts~\citep{yu2026w2sopd,feng2026directopd,park2026oprd}, but do not systematically characterize how much and how fast capability transfers across scales.

In this work, we study the scaling properties of OPD by asking three questions.
\textbf{First}, do OPD training dynamics contain a regular, predictable regime, and how long does it last?
\textbf{Second}, how do student and teacher scale shape peak gold score, useful-transfer rate, and transfer extent, and can the fitted laws predict outcomes at held-out scales?
\textbf{Third}, how do design choices such as the OPD objective (\textit{Vanilla-OPD} vs.\ \textit{Delta-OPD}) and the degree of on-policy supervision change transfer, and does bootstrapping weak-to-strong OPD along a model family improve on direct transfer from the smallest expert?

We investigate these questions via controlled experiments on math reasoning with Qwen2.5 models (0.5B--14B), fitting power laws in student and teacher parameter count and measured teacher score for peak gold score and useful-transfer slope (\cref{fig:overview}).
Our \textbf{main findings}:
\begin{itemize}[leftmargin=*,topsep=0pt,itemsep=2pt,parsep=0pt]
\item \textbf{Capability transfer has a regular initial regime and heterogeneous later dynamics.}
Initial OPD training dynamics exhibit a regular useful-transfer regime, with gold score increasing linearly with training progress ($d$) (\cref{sec:dynamics}).

\item \textbf{Scale predicts peak capability and local useful-transfer rate.}
Peak remaining error and useful-transfer rate follow joint power laws in student size, effective teacher size, and measured teacher score, with the peak law extrapolating to the largest held-out scales within one accuracy point.
At a matched score, smaller teachers transfer better (\cref{sec:scale_laws}).

\item \textbf{The distillation objective changes capability transfer.}
Delta-OPD produces a larger local matched-KL gain slope in 15 shared pairs and a larger observed peak gain in 12, with the advantage concentrated in weak-to-strong pairs (\cref{sec:extensions}).

\item \textbf{Off-policy cold start harms weak-to-strong OPD.}
An off-policy SFT phase harms OPD increasingly with the weak-to-strong capability gap (\cref{sec:onpolicyness}).

\item \textbf{Bootstrapping weak-to-strong OPD does not improve on direct transfer.}
Every bootstrapped chain peaks below direct OPD from the smallest post-RL expert at the same student scale, even though its intermediate teachers attain higher gold scores (\cref{sec:bootstrap}).
\end{itemize}

%%%%%%%%%%%%%%%%%%%%%%%%%%%%%%%%%%%%%%%%%%%%%%%%%%%%%%%%%%%%%%%%%%%%%%%%%%%%%%%
%%%%%%%%%%%%%%%%%%%%%%%%%%%%%%%%%%%%%%%%%%%%%%%%%%%%%%%%%%%%%%%%%%%%%%%%%%%%%%%
\section{Preliminaries}
\label{sec:preliminary}

This section fixes notation and formalizes the three distillation objectives used throughout the paper.

\mypar{Vanilla-OPD.}
Let $x\!\sim\!\gD$ be a prompt, $y$ a student rollout, $\teacher$ a fixed teacher, and $\policy$ the policy parameterized by $\theta$.
The original OPD recipe minimizes the sequence-level reverse KL divergence between the policy and the teacher~\citep{agarwal2024onpolicy,gu2024minillm} (subscript V for ``Vanilla''):
\begin{equation} \label{eq:sequence_opd}
\min_\theta \gL_{\mathrm{V}}(\theta)
 =\min_\theta \underset{\substack{x\sim\gD,\,y\sim\policy(\cdot\vert x)}}{\E}\left[
 \KL\!\left(
 \policy(y\vert x)\Vert\teacher(y\vert x)\right)
 \right].
\end{equation}
By the chain rule of KL divergence, minimizing $\gL_{\mathrm V}$ is equivalent to maximizing an RL-like objective in which token $t$ receives the teacher-induced token reward $r_t=\log\teacher(y_t\vert x,y_{<t})-\log\policy(y_t\vert x,y_{<t})$, and the exact policy gradient assigns token $t$ the return-to-go advantage $\smash{{\sum\nolimits}_{t'=t}^{|y|} r_{t'}}$.
Modern implementations instead apply a \textit{zero-discount update}, discounting future rewards to zero so that each token receives only its immediate reward, $A_t^{\mathrm V}=r_t$~\citep{lu2025onpolicydistillation}.
This practice is widely adopted in frontier post-training pipelines~\citep{yang2025qwen3,xiao2026mimo,zeng2026glm}:
\begin{equation} \label{eq:opd}
\max_\theta \gJ_{\mathrm V}(\theta)
=\max_\theta \underset{\substack{x\sim\gD,\,y\sim\policy(\cdot\vert x)}}{\E} \left[
{\sum\nolimits}_{t=1}^{|y|}
\underbrace{\log\teacher(y_t\vert x,y_{<t}) -\log\policy(y_t\vert x,y_{<t})}_{A_t^\mathrm{V}(x,y)}
\right].
\end{equation}
In expectation, $-A_t^{\mathrm V}$ equals the full-vocabulary conditional reverse KL at each student-visited prefix.
Estimating this divergence with only the sampled token's logprobs, rather than a sum over the full vocabulary, is known as \textit{sampled-token estimation}~\citep{li2026rethinking}, which we adopt for Vanilla-OPD under the policy-gradient framework by default.

\mypar{Delta-OPD.}
The second variant, which we call Delta-OPD, derives its token reward from the policy shift the teacher acquired during RL.
This reward construction is the common core of variants proposed for both transfer directions: OPD$^2$~\citep{heo2026opd2} applies it to strong-to-weak distillation, and Direct-OPD~\citep{feng2026directopd} and W2S-OPD~\citep{yu2026w2sopd} apply it to weak-to-strong distillation.
We therefore adopt it as the representative alternative objective for a design that spans weak-to-strong, same-base, and strong-to-weak pairs.
Let $\teacherbase$ be the SFT checkpoint from which $\teacher$ was produced by RL, and $\reference$ be the student base policy.
The objective is as follows (superscript $\Delta$ for ``Delta''):
\begin{equation} \label{eq:delta}
\begin{gathered}
A_t^\Delta(x,y)
= r_t^\Delta
\coloneqq \log\teacher(y_t\vert x,y_{<t})
  -\log\teacherbase(y_t\vert x,y_{<t}), \\
\max_\theta \gJ_\Delta(\theta)
=\max_\theta \underset{\substack{x\sim\gD,\,y\sim\policy(\cdot\vert x)}}{\E} \left[
{\sum\nolimits}_{t=1}^{|y|}
A_t^\Delta(x,y)
-
\KL\!\left(\policy(\cdot\vert x,y_{<t})\Vert
\reference(\cdot\vert x,y_{<t})\right)
\right].
\end{gathered}
\end{equation}

\mypar{Off-policy distillation (OffPD).}
When rollouts are instead sampled from the teacher and the student maximizes their likelihood, as in SFT on teacher demonstrations, distillation becomes sequence-level \textit{off-policy distillation (OffPD)}~\citep{hinton2015distilling,kim2016sequence}:
\begin{equation}
\min_{\theta} \gL_{\mathrm{Off}}(\theta)
= \min_\theta \underset{x\sim\gD,\,y\sim\teacher(\cdot\vert x)}{\E}\left[
  -\log \policy(y \vert x)
\right].
\end{equation}

%%%%%%%%%%%%%%%%%%%%%%%%%%%%%%%%%%%%%%%%%%%%%%%%%%%%%%%%%%%%%%%%%%%%%%%%%%%%%%%
%%%%%%%%%%%%%%%%%%%%%%%%%%%%%%%%%%%%%%%%%%%%%%%%%%%%%%%%%%%%%%%%%%%%%%%%%%%%%%%
\section{Experimental design}
\label{sec:setup}

\mypar{Models and training pipeline.}
The study uses Qwen2.5 Base (not instruction-tuned) models~\citep{qwen2024qwen25} at 0.5B, 1.5B, 3B, 7B, and 14B parameters.
All models undergo an initial SFT phase on a subset of Dolci-SFT~\citep{olmo2025olmo} for basic instruction-following under a chat template.
Teacher models are obtained by GRPO RL on the mixed GSM8K~\citep{cobbe2021gsm8k} and MATH~\citep{hendrycksmath2021} training split of 14.8K examples and evaluated on the corresponding mixed test split of 6.3K examples (results in \cref{fig:rl_trajectories}).
OPD and RL use the same training prompts, and each OPD run lasts at most ten epochs (580 updates) with periodic held-out evaluation.
All training phases are implemented in the verl framework~\citep{sheng2025hybridflow} (hyperparameters in \cref{app:hyperparameters}).
The design contains 25 teacher--student combinations, including weak-to-strong, same-base, and strong-to-weak setups.

\mypar{Metrics.}
Gold score is accuracy on the held-out test set, on which both the RL teachers and the OPD students are evaluated.
It plays the role of the gold reward in~\citet{gao2023scaling}, with the teacher-induced token reward as the proxy.
For prompts $x\!\sim\!\gD$ and student rollouts $y$, define
\begin{equation} \label{eq:k3}
\begin{gathered}
k_{3}(\policy, \reference)
=\underset{x \sim \gD, y \sim \policy(\cdot|x)}{\E}\left[
  \frac{1}{|y|}{\sum\nolimits}_{t=1}^{|y|}
  \exp(\delta_{t})-\delta_{t}-1
\right],
  \\
\text{where}~ \delta_{t}=\log\reference(y_{t}\vert x,y_{<t})
-\log\policy(y_{t}\vert x,y_{<t}).
\end{gathered}
\end{equation}
This is the nonnegative, unbiased $k_3$ estimator of reverse KL~\citep{schulman2020kl,tang2025few}, averaged over response tokens.
Unlike \citet{gao2023scaling}, who measure sequence-level KL, we measure token-mean KL as this matches the immediate-token objective in \cref{eq:opd}.
Since KL is a quadratic measure, we use its square root, $d\!\coloneqq\!\smash{\sqrt{k_{3}}}$, as a proxy for training progress.

%%%%%%%%%%%%%%%%%%%%%%%%%%%%%%%%%%%%%%%%%%%%%%%%%%%%%%%%%%%%%%%%%%%%%%%%%%%%%%%
%%%%%%%%%%%%%%%%%%%%%%%%%%%%%%%%%%%%%%%%%%%%%%%%%%%%%%%%%%%%%%%%%%%%%%%%%%%%%%%
\section{Characterizing capability-transfer dynamics in OPD}
\label{sec:dynamics}

This section addresses our first question: \textit{do OPD training dynamics contain a regular, predictable regime, and how long does it last?}
We examine the checkpoint trajectories of the 25 Vanilla-OPD runs, which span weak-to-strong, same-base, and strong-to-weak setups, reading each run as a curve of gold score $G$ against training progress $d$ (\cref{sec:setup}) following \citet{gao2023scaling}.
The quantities introduced here, namely the transfer rate, the transfer extent, and the peak, are what \cref{sec:scale_laws} later characterizes across teacher and student scale.

\begin{figure}[t]
\centering
\includegraphics[width=\linewidth,keepaspectratio]{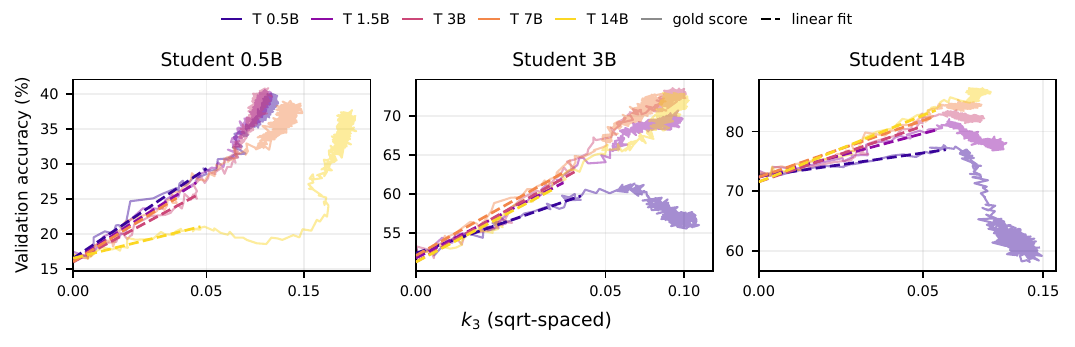}
\caption{Gold score for the 0.5B, 3B, and 14B students as teacher scale varies.
Dashed lines show linear fits to the first 30 observations.
Full results are in \cref{fig:all_curves,fig:delta_all_curves}.}
\label{fig:representative}
\end{figure}

Two phases emerge consistently.
Transfer begins with a regular useful-transfer regime, in which gold score rises approximately linearly in $d$ at a rate $m$.
We define the \textit{transfer endpoint} $d_{\mathrm{transfer}}$ as the last $d$ before the trajectory falls below the 95\% predictive band of its initial linear fit for three consecutive checkpoints.
Beyond $d_{\mathrm{transfer}}$, dynamics are noisy and heterogeneous, mixing attenuated improvement, saturation, and regression around the gold-score maximum at $d_{\mathrm{peak}}$.
\Cref{fig:representative} overlays all teachers for three representative students, and \cref{fig:all_curves,fig:delta_all_curves} show every pair.
Every trajectory improves initially, and peak gold score increases with teacher scale for the larger students, from 72.7\% to 81.9\% for the 7B student and from 77.7\% to 87.3\% for the 14B student.
In every panel the smallest teacher shows the clearest post-peak regression, while other tails attenuate or saturate.
For smaller students, larger teachers do not always improve the peak: the
0.5B student peaks at 40.8\% with the 3B teacher but reaches only 39.0\%
under the 7B teacher and 37.7\% under the 14B teacher.

\begin{wrapfigure}{r}{0.38\linewidth}
\centering
\includegraphics[width=\linewidth]{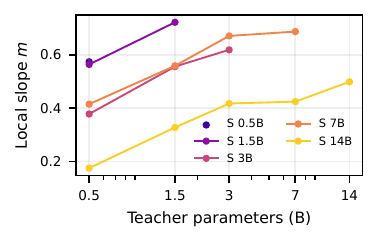}
\caption{Fitted initial slope $m$ across teacher scale, including only
teachers no larger than each fixed student.}
\label{fig:transfer_rate}
\end{wrapfigure}
\mypar{Initial transfer is approximately linear in the KL coordinate.}
Across the 25 runs, linear fits $G(d)=c+md$ to the first 30 checkpoints obtain
$R^2\in[0.932,0.988]$ and RMSE in $[0.0026,0.0129]$ accuracy units.  The fitted
slopes span $[0.175,0.721]$: weak teachers generally induce smaller initial
gains for larger students, while teachers closer to the student's scale tend to
raise gold score faster per unit $d$ (\cref{fig:transfer_rate}).
The observed tails include attenuated improvement, near-saturation, and
regression, with no shared functional form.

\mypar{Why square-root KL linearizes local transfer.}
Local KL geometry explains the observed linearity: along a smooth training path, gold score changes to first order in the parameter perturbation while KL changes to second order, so $G(d)=G(0)+md+O(d^2)$ with a slope set by the Fisher-normalized alignment of the update direction with the gold-score gradient (statement and proof in \cref{app:local_geometry}).

\subsection{How scale shapes transfer and later dynamics}
\label{sec:scaling}

\begin{figure}[t]
\centering
\includegraphics[width=\linewidth]{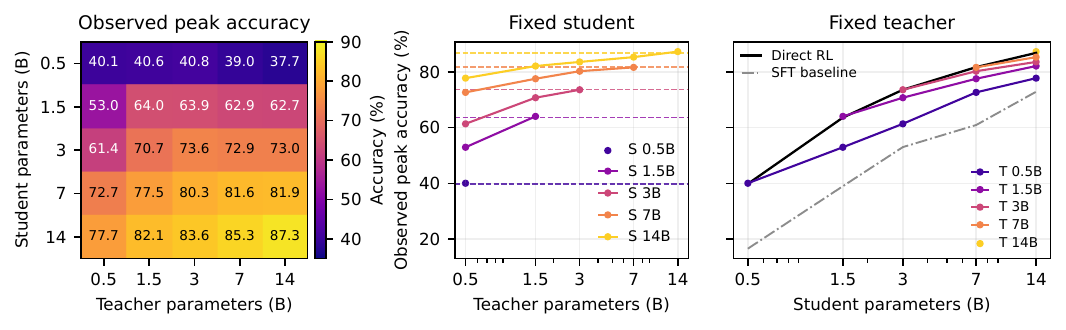}
\caption{Peak gold score across model scale.
Left: the complete 25-cell Vanilla grid.
Middle: fixed-student Vanilla trends using teachers no larger than the student,
with color-matched dashed levels marking each student's direct-RL endpoint.
Right: Vanilla peaks grouped by teacher, together with the direct-RL endpoint
and the SFT baseline, on the same y axis as the middle panel.
Parameter axes are logarithmic and all curves are one-run descriptions.}
\label{fig:scaling}
\end{figure}

\Cref{fig:scaling} relates the complete Vanilla-OPD grid to direct student RL.
At fixed teacher size, observed peak gold score increases with student scale.
At fixed 7B and 14B students, it also increases monotonically with teacher size.
The fixed-3B-student series instead peaks with the 3B teacher rather than the
7B teacher, and the 0.5B and 1.5B students lose accuracy under their largest
teachers.
Vanilla-OPD same-base peaks lie within 0.5 percentage points of the final direct-RL
reference at all five scales, exceeding it at three of them.

The locations $d_{\mathrm{peak}}$ of the gold-score maximum are less monotone: for the 7B student they are 0.274, 0.282, 0.308, 0.323, and 0.323 as teacher size increases, and other student series likewise contain reversals.
We therefore track the peak value $G_{\mathrm{peak}}$, initial rate $m$, and
transfer extent $d_{\mathrm{transfer}}$ rather than imposing a parametric law
on the location of the observed maximum.

\mypar{Does the teacher proxy outlive the gold gain?}
The teacher-induced implicit reward yields a continuous proxy score $P$, logged as the token-mean reward on training rollouts.
Whether $P$ keeps improving after gold score peaks distinguishes implicit-reward overoptimization from simple over-imitation.
\Cref{fig:all_curves} overlays the logged proxy curves with gold score: wherever gold score regresses, $P$ keeps rising, so late-stage regression carries the signature of implicit-reward overoptimization.

\mypar{When does the initial law cease to predict?}
The transfer endpoint is a sequential predictive quantity rather than a fitted
turning point.
Under the primary 95\% band and three-consecutive-deviation rule, nine of 25
Vanilla-OPD trajectories depart within their observed support.
Changing the confidence level and persistence requirement varies this count
only from eight to twelve (\cref{tab:endpoint_sensitivity}).

%%%%%%%%%%%%%%%%%%%%%%%%%%%%%%%%%%%%%%%%%%%%%%%%%%%%%%%%%%%%%%%%%%%%%%%%%%%%%%%
%%%%%%%%%%%%%%%%%%%%%%%%%%%%%%%%%%%%%%%%%%%%%%%%%%%%%%%%%%%%%%%%%%%%%%%%%%%%%%%
\section{Fitting power laws for peak gold score and useful-transfer rate}
\label{sec:scale_laws}

This section addresses our second question: \textit{how do student and teacher scale shape peak gold score, useful-transfer rate, and transfer extent, and do the fitted laws extrapolate to held-out scales?}
Since dynamics are regular only up to the transfer endpoint (\cref{sec:dynamics}), we fit the rate $m$ and the peak $G_{\mathrm{peak}}$ as separate targets and evaluate every law by withholding the largest models, while the extent $d_{\mathrm{transfer}}$ is summarized as an empirical KL budget.

We normalize parameter counts by one billion and cap teacher scale at student
scale,
$\widetilde N_S=N_S/(1\mathrm B)$ and
$\widetilde N_T^{\mathrm{eff}}=\min(N_T,N_S)/(1\mathrm B)$.
The cap approximates the saturation and reversal observed above student scale (\cref{fig:scaling}); the smallest students' reversals remain unmodeled residuals.

\mypar{Joint laws in scale and teacher score.}
Parameter count summarizes a teacher only when every teacher is trained to its RL endpoint.
An undertrained teacher scores below the size trend.
Our primary model therefore conditions the peak and rate on the remaining error of the effective teacher, whose gold score is $G_T^{\mathrm{eff}}$, fitting Vanilla-OPD and Delta-OPD independently with the shared functional families
\begin{equation}
1-G_{\mathrm{peak}}
=A\widetilde N_S^{-\alpha}
(\widetilde N_T^{\mathrm{eff}})^{-\beta}
\left(1-G_T^{\mathrm{eff}}\right)^{\zeta},
\qquad
m
=B\widetilde N_S^{-\gamma}
(\widetilde N_T^{\mathrm{eff}})^{\delta}
\left(1-G_T^{\mathrm{eff}}\right)^{-\xi}.
\label{eq:teacher_score_scale}
\end{equation}
On the grid whose teachers are all RL endpoints, the two covariates are collinear ($r=-0.9999$), so the exponents of \cref{eq:teacher_score_scale} cannot be separated there.
The bootstrapped chains of \cref{sec:bootstrap} break this collinearity with teachers that are themselves OPD products of a previous chain stage, whose measured gold scores and initial slopes sit below the size trend.
Adding their five Vanilla-OPD and three Delta-OPD cells lowers the collinearity to $-0.95$ and $-0.97$ and identifies every teacher exponent, with all bootstrap intervals excluding zero (\cref{app:scale_laws}).
\Cref{tab:scaling_coefficients} reports the fitted joint laws.

\mypar{Motivation for multiplicative power law.}
Scaling the student removes a constant fraction of whatever error the teacher-induced supervision leaves, and the multiplicative family encodes exactly this interaction.
An additive alternative instead posits a teacher-induced error that persists as $N_S \to \infty$ and yields higher fitting errors (\cref{app:scale_laws}).

\mypar{Scale-only baseline.}
Dropping the teacher-score factor yields the family that reads scaling off parameter counts alone,
\begin{equation}
1-G_{\mathrm{peak}}
=A\widetilde N_S^{-\alpha}
(\widetilde N_T^{\mathrm{eff}})^{-\beta},
\qquad
m
=B\widetilde N_S^{-\gamma}
(\widetilde N_T^{\mathrm{eff}})^{\delta}.
\label{eq:peak_scale}
\end{equation}
This family is the baseline that \cref{tab:peak_validation} compares the joint laws against.

\begin{remark}[On parameter count as the scaling variable]
Parameter count is a confounded measure that implicitly incorporates model scale, data scale, and training compute.
Since \citet{qwen2024qwen25} tuned architecture and data scale by scaling laws, our power laws should be read~at~\mbox{compute-optimal}~settings.
\end{remark}

\begin{table}[t]
\caption{Fitted joint power-law coefficients (\cref{eq:teacher_score_scale}).}
\label{tab:scaling_coefficients}
\footnotesize
\begin{subtable}[t]{0.48\linewidth}
\caption{Peak capability (\cref{eq:teacher_score_scale}, left).}
\label{tab:peak_coefficients}
\centering
\begin{tabular}{lcccc}
\toprule
OPD method & $A$ & $\alpha$ & $\beta$ & $\zeta$ \\
\midrule
Vanilla-OPD & 0.97 & 0.30 & $-0.27$ & 0.95 \\
Delta-OPD & 1.02 & 0.34 & $-0.33$ & 1.01 \\
\bottomrule
\end{tabular}
\end{subtable}
\hfill
\begin{subtable}[t]{0.48\linewidth}
\caption{Useful-transfer rate (\cref{eq:teacher_score_scale}, right).}
\label{tab:rate_coefficients}
\centering
\begin{tabular}{lcccc}
\toprule
OPD method & $B$ & $\gamma$ & $\delta$ & $\xi$ \\
\midrule
Vanilla-OPD & 0.12 & 0.19 & $-0.61$ & 1.90 \\
Delta-OPD & 0.13 & 0.20 & $-0.73$ & 2.01 \\
\bottomrule
\end{tabular}
\end{subtable}
\end{table}

\begin{table}[t]
\caption{Validation of the peak laws, in accuracy points.
Leave-one-scale-out RMSE refits each law with all cells sharing one student or teacher scale
withheld and predicts the withheld cells (mean over the two split types); extrapolation RMSE
withholds the largest student (S) or teacher (T) scale.
Protocols and further diagnostics are in \cref{app:scale_laws}.
Best results are highlighted in bold.}
\label{tab:peak_validation}
\centering
\footnotesize
\begin{tabular}{llcc}
\toprule
OPD method & Peak law & Leave-one-scale-out ($\downarrow$) & Extrapolation S / T ($\downarrow$) \\
\midrule
\multirow{3}{*}{Vanilla-OPD}
 & Scales only (\cref{eq:peak_scale}) & 3.43 & .64 / .75 \\
 & Teacher score only & 2.55 & 2.54 / .70 \\
 & Joint (\cref{eq:teacher_score_scale}) & \textbf{1.66} & \textbf{.55} / \textbf{.68} \\
\midrule
\multirow{3}{*}{Delta-OPD}
 & Scales only (\cref{eq:peak_scale}) & 2.47 & .21 / \textbf{.29} \\
 & Teacher score only & 2.07 & .93 / .46 \\
 & Joint (\cref{eq:teacher_score_scale}) & \textbf{0.82} & \textbf{.20} / .32 \\
\bottomrule
\end{tabular}
\end{table}

\begin{figure}[t]
\centering
\includegraphics[width=.52\linewidth]{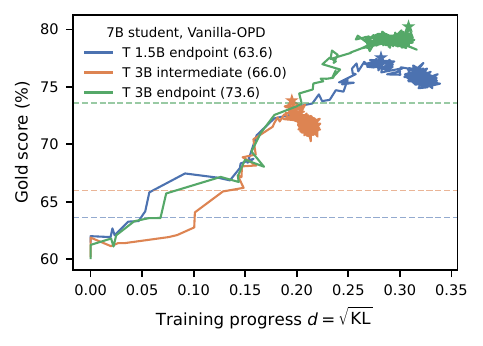}
\caption{Trajectories of the intermediate-teacher validation.
Gold score against $d$ for the 7B student distilled from
the 1.5B RL endpoint, the score-matched intermediate 3B checkpoint at step
58, and the 3B RL endpoint.
Stars mark peaks and dashed levels each teacher's own gold score.}
\label{fig:intermediate_teacher}
\end{figure}

\mypar{Peak capability.}
In every observed weak-to-strong pair, the student's peak gold score exceeds its teacher's own, by margins that shrink as teacher scale approaches student scale (\cref{fig:scaling}).
With $\zeta\approx1$ for both methods, peak student error is nearly proportional to the capped teacher's remaining error, scaled down by a power of student size; the negative $\beta$ means that at matched score the smaller teacher transfers better.
\Cref{tab:peak_validation} validates these laws: conditioning on teacher score halves the leave-one-scale-out RMSE of the scale-only baseline of \cref{eq:peak_scale}, and the joint law extrapolates to the held-out largest student and teacher within 0.7 (Vanilla-OPD) and 0.4 (Delta-OPD) accuracy points.
A controlled comparison tests the negative $\beta$ out of sample: we distill the 7B student from an intermediate 3B teacher checkpoint scoring 66.0, slightly above the 1.5B RL endpoint's 63.6 (\cref{tab:intermediate_teacher,fig:intermediate_teacher}).
The joint law predicts its peak within 0.3 points (74.1 against the observed 73.8) and the correct ordering below the 1.5B teacher's 77.5.
The scale-only and score-only laws instead both favor the larger, higher-scoring teacher; protocol details are in \cref{app:scale_laws}.

\begin{table}[t]
\caption{Out-of-sample validation of the Vanilla-OPD peak laws with an
intermediate 3B teacher checkpoint, in gold-score points.
Each row is a 7B-student cell; predictions use the full-fit coefficients of
the scale-only (\cref{eq:peak_scale}), score-only
(\cref{eq:perf_peak_scale}), and joint (\cref{eq:teacher_score_scale}) laws.
The intermediate-teacher cell is held out from every fit; only the joint law
orders it correctly relative to the 1.5B endpoint teacher.}
\label{tab:intermediate_teacher}
\centering
\footnotesize
\begin{tabular}{lccccc}
\toprule
Teacher & Teacher score & Observed peak & Scales only & Score only & Joint \\
\midrule
1.5B endpoint (step 580) & 63.6 & 77.5 & 77.0 & 75.6 & \textbf{77.1} \\
3B intermediate (step 58) & 66.0 & 73.8 & 79.5 & 76.3 & \textbf{74.1} \\
3B endpoint (step 580) & 73.6 & 80.3 & 79.5 & 78.7 & \textbf{79.6} \\
\bottomrule
\end{tabular}
\end{table}

\mypar{Useful-transfer rate.}
The rate law mirrors the peak law with stronger score dependence: $\xi=1.9$ for Vanilla-OPD and $2.0$ for Delta-OPD, so transfer slows sharply as the teacher's remaining error grows, and at matched score the larger teacher transfers more slowly.
Rate fits are noisier than peak fits (log-space $R^2$ of 0.59 and 0.76); the rate law is better read as an interpretable scale summary than a precise predictor (\cref{app:scale_laws}).

\begin{figure}[t]
\centering
\begin{subfigure}[t]{0.49\linewidth}
\includegraphics[width=\linewidth]{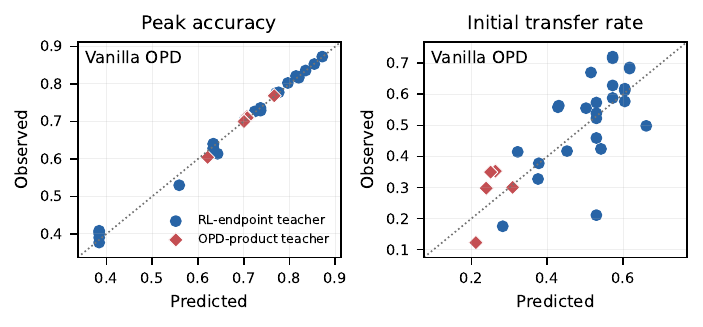}
\caption{Vanilla-OPD}
\label{fig:scale_diagnostics}
\end{subfigure}
\hfill
\begin{subfigure}[t]{0.49\linewidth}
\includegraphics[width=\linewidth]{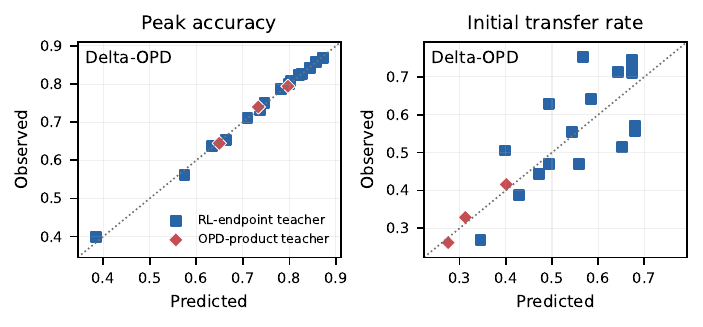}
\caption{Delta-OPD}
\label{fig:scale_diagnostics_delta}
\end{subfigure}
\caption{Observed peak gold score and initial transfer rate against full-fit
predictions of the joint laws of \cref{eq:teacher_score_scale}.
Red diamonds mark cells whose teachers are OPD products of the bootstrapped
chains (\cref{sec:bootstrap}).}
\label{fig:scale_diagnostics_joint}
\end{figure}

\mypar{Useful-transfer extent.}
The transfer extent resists a comparable law.
Observed departures and censored lower bounds together span $d$ of 0.20--0.36 for Vanilla-OPD and 0.27--0.34 for Delta-OPD, within a factor of two across the grid, with medians of 0.30 and 0.29.
A censored accelerated-failure-time fit finds only weak scale dependence, and its Delta-OPD exponents are not identifiable (\cref{app:scale_laws}).
We therefore read the extent as an approximately scale-free KL budget rather than a scaling target.
\Cref{fig:scale_diagnostics,fig:scale_diagnostics_delta} plot predicted against observed quantities for both methods; the extent diagnostics are in \cref{app:scale_laws}.

%%%%%%%%%%%%%%%%%%%%%%%%%%%%%%%%%%%%%%%%%%%%%%%%%%%%%%%%%%%%%%%%%%%%%%%%%%%%%%%
%%%%%%%%%%%%%%%%%%%%%%%%%%%%%%%%%%%%%%%%%%%%%%%%%%%%%%%%%%%%%%%%%%%%%%%%%%%%%%%
\section{Effects of design choices on OPD transfer}
\label{sec:design}

This section addresses our third question: \textit{how do design choices change OPD transfer?}
We vary one choice at a time: the OPD objective, the degree of on-policy supervision, and bootstrapping.

\subsection{Effect of OPD variant: Delta-OPD}
\label{sec:extensions}

Our analyses above focus on Vanilla-OPD.
Since many OPD variants have appeared recently, we take Delta-OPD (\cref{sec:preliminary}) as the alternative condition and ask how the objective changes transfer.

We compare 17 Delta-OPD runs with Vanilla-OPD cells having the same teacher and
student scales.
Ten pairs are weak-to-strong, five are same-base, and
0.5B$\leftarrow$1.5B and 1.5B$\leftarrow$3B are strong-to-weak controls.
Because separately sampled step-zero accuracies differ, the primary quantity is gain over each run's own initialization, $\smash{\Delta G_{\mathrm V}(d)=G_{\mathrm V}(d)-G_{\mathrm V}(0)}$ and $\smash{\Delta G_\Delta(d)=G_\Delta(d)-G_\Delta(0)}$.
The local slope compares matched-KL gold gain.
We estimate it from 30 Vanilla-OPD and 40 Delta-OPD observations (Delta-OPD's regular phase spans more checkpoints) and restrict fitted comparisons to their common $d$ support.

\begin{wrapfigure}{r}{0.45\linewidth}
\centering
\includegraphics[width=\linewidth]{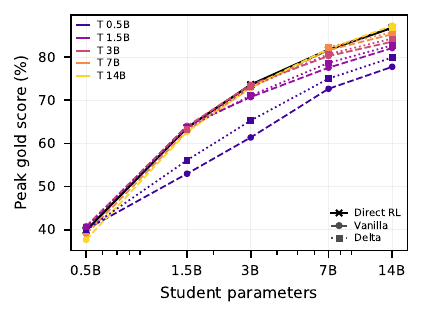}
\caption{Peak gold score against student scale for Delta-OPD, Vanilla-OPD,
and direct RL, grouped by teacher.}
\label{fig:direct}
\end{wrapfigure}
First-40 Delta-OPD lines obtain $R^2\in[0.916,0.987]$, RMSE in $[0.0049,0.0116]$ (\cref{fig:delta_all_curves}).
Their slopes exceed Vanilla-OPD in 15 of the 17 shared cells; the exceptions are the two smallest students under the 0.5B teacher (\cref{tab:delta_full}).
Delta-OPD attains the larger baseline-normalized peak gain and the larger absolute peak in 12 cells each, mostly in weak-to-strong pairs (nine of ten; \cref{fig:direct}).
The benefit shrinks with teacher scale: the 0.5B teacher yields two to four extra points of absolute peak, whereas larger teachers stay within about one point.
Five of the 17 runs depart from the initial line.

\subsection{Degree of on-policy supervision}
\label{sec:onpolicyness}
Our previous results are based on pure OPD, where a teacher policy directly supervises a base model.
Since SFT on teacher rollouts is often used as a cold start that aligns the student's output distribution with the teacher's~\citep{li2026rethinking}, we compare three degrees of on-policy supervision: purely on-policy, on-policy with off-policy cold start, and purely off-policy.

\begin{wrapfigure}{r}{0.45\linewidth}
\centering
\includegraphics[width=\linewidth]{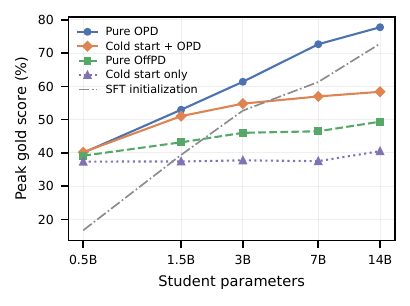}
\caption{Peak gold score under three degrees of on-policy supervision (0.5B
expert as teacher).}
\label{fig:onpolicyness}
\end{wrapfigure}
The three conditions share teachers and the evaluation protocol of \cref{sec:setup}.
Pure OPD trains as before.
The cold-start condition first performs one epoch of SFT on teacher rollouts over 20K Dolci-Think-RL prompts~\citep{olmo2025olmo}, then runs the full 580 OPD updates.
Pure OffPD trains on teacher rollouts alone, five sampled responses per training prompt for five epochs (\cref{sec:preliminary}).
We report each condition's peak gold score over its training budget, with the 0.5B expert as teacher in \cref{fig:onpolicyness} and the full weak-to-strong and same-base grid in \cref{fig:onpolicyness_grid}.

Pure OPD attains the best peak gold score in every cell (\cref{fig:onpolicyness}).
The off-policy cold start grows increasingly harmful with the weak-to-strong capability gap, costing 6.6, 15.7, and 19.4 points for the 3B, 7B, and 14B students taught by the 0.5B expert.
The damage is done by the cold start itself: one epoch of SFT on the 0.5B expert's rollouts pins every student near the teacher's own score, erasing 32 points of the 14B student's initialization.
The subsequent OPD phase does not recover the loss (dynamics in \cref{app:extensions}).
Pure OffPD underperforms pure OPD in every cell, by 28.4 points in the extreme weak-to-strong cell and by less than one point in same-base cells (\cref{fig:onpolicyness_grid}), so on-policy supervision matters most where weak-to-strong transfer is most attractive.

\subsection{Bootstrapping weak-to-strong OPD}
\label{sec:bootstrap}

\citet{burns2024weak} report that bootstrapping weak-to-strong fine-tuning through intermediate model sizes can outperform a single direct step in their chess puzzle setting, and we test whether the same holds for OPD.
The preceding experiments obtain a teacher at scale $X$ by RL on the $X$-scale SFT model.
Here RL is applied only to the smallest (0.5B) model, and every larger teacher in the chain is the previous stage's best OPD checkpoint:
$\gM_1^{\mathrm{SFT}} \xrightarrow{{\mathrm{RL}}} \gM_1 \xrightarrow{{\mathrm{OPD}}} \gM_2 \xrightarrow{{\mathrm{OPD}}} \cdots \xrightarrow{{\mathrm{OPD}}} \gM_n$.
We run five Vanilla-OPD chains and three Delta-OPD chains, stepping through every intermediate size (up to 0.5B$\to$1.5B$\to$3B$\to$7B$\to$14B) or skipping sizes (e.g.\ 0.5B$\to$1.5B$\to$7B).

\begin{wrapfigure}{r}{0.40\linewidth}
\centering
\includegraphics[width=\linewidth]{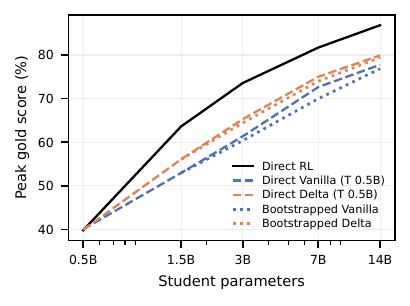}
\caption{Full 0.5B$\to$14B bootstrapped chains against direct baselines.}
\label{fig:bootstrap}
\end{wrapfigure}
We compare each bootstrapped student against direct OPD from the 0.5B expert at the same scale, with direct RL endpoints as references (\cref{fig:bootstrap}).
Bootstrapping does not improve on direct transfer: every chain peaks below direct OPD from the 0.5B expert (60.4 versus 61.4 at 3B, 70.0--71.3 versus 72.7 at 7B, and 76.8 versus 77.7 at 14B, in \%), and longer chains end slightly lower than shorter ones.
The comparison also dissociates a teacher's score from its teaching value: the 1.5B OPD product scores 53.0\% against the 0.5B RL expert's 39.8\%, yet its 3B student peaks below the 0.5B expert's direct student.

This reverses the bootstrapping gains of \citet{burns2024weak} and echoes \citet{li2026rethinking}: a higher-scoring teacher helps only when it offers new capabilities.
A candidate explanation is the feature-slots picture of \citet{huang2026larger}, in which each OPD stage bottlenecks the features inherited from the RL expert.
The three Delta-OPD chains behave alike, the full one peaking below direct Delta-OPD from the 0.5B expert at every stage (64.5 versus 65.3 at 3B, 74.0 versus 75.0 at 7B, and 79.4 versus 79.9 at 14B).
Direct RL on the student itself remains above every weak-teacher variant, so the appeal of weak-to-strong OPD rests on amortizing one expert across a model family rather than on beating direct RL (per-chain dynamics in \cref{app:extensions}).

%%%%%%%%%%%%%%%%%%%%%%%%%%%%%%%%%%%%%%%%%%%%%%%%%%%%%%%%%%%%%%%%%%%%%%%%%%%%%%%
%%%%%%%%%%%%%%%%%%%%%%%%%%%%%%%%%%%%%%%%%%%%%%%%%%%%%%%%%%%%%%%%%%%%%%%%%%%%%%%
\section{Related work}
\label{sec:related}

\mypar{On-policy distillation.}
In OPD, a teacher policy provides token-level supervision on rollouts sampled from the student, avoiding the exposure bias of offline teacher data~\citep{agarwal2024onpolicy,gu2024minillm,lu2025onpolicydistillation}.
Variants reshape the token reward and objective~\citep{song2026survey,yang2026learning,ko2026relaxed}, and empirical analyses dissect failure modes and recipes~\citep{fu2026revisiting,li2026rethinking}.
Closest to us, OPD$^2$~\citep{heo2026opd2}, Direct-OPD~\citep{feng2026directopd}, and W2S-OPD~\citep{yu2026w2sopd} supervise the student with contrasts between post-RL and pre-RL teacher checkpoints, a family our Delta-OPD instantiates, and OPRD instead rescales the student's verifier-driven RL gradient along this shift~\citep{park2026oprd}.
These works design and diagnose objectives; we characterize how fixed objectives scale.

\mypar{Reward model overoptimization.}
\citet{gao2023scaling} model gold reward as a function of KL from the initial policy with coefficients predicted from reward-model scale, a roadmap extended to direct alignment~\citep{rafailov2024scaling} and surveyed as reward hacking~\citep{wang2026reward}.
Distillation inherits this proxy-optimization hazard: repeatedly fitting fixed offline teacher data induces teacher hacking~\citep{tiapkin2025on}, while noisy-expert theory favors online interaction~\citep{sriraman2026behavior}.
The teacher-induced token reward is likewise a gold-score proxy, motivating our analysis.

\mypar{Weak-to-strong generalization.}
Weak supervision can elicit capabilities beyond the supervisor's own~\citep{burns2024weak,ildiz2024highdim,yuan2026incentivizing}, with feature-learning theory attributing the gains to knowledge latent in the strong model~\citep{awano2026feature}.
In pretraining distillation, small teachers also improve larger students, with gains saturating or reversing as teachers grow~\citep{lu2026strong}.
Our peak scaling law quantifies this regime via the effective-teacher cap.

\mypar{Predicting model performance with scaling laws.}
Scaling laws first predicted pretraining loss from model size, data, and compute~\citep{kaplan2020scaling,hoffmann2022training} and have since been fit for fine-tuning~\citep{zhang2024finetuning}, pretraining distillation~\citep{busbridge2025distillation}, and RL compute~\citep{khatri2026art}, and the pretrained state predicts later RL gains~\citep{shen2026reasoning}.
These laws predict endpoints from resource budgets, while ours target KL-indexed trajectories.

%%%%%%%%%%%%%%%%%%%%%%%%%%%%%%%%%%%%%%%%%%%%%%%%%%%%%%%%%%%%%%%%%%%%%%%%%%%%%%%
%%%%%%%%%%%%%%%%%%%%%%%%%%%%%%%%%%%%%%%%%%%%%%%%%%%%%%%%%%%%%%%%%%%%%%%%%%%%%%%
\section{Conclusion}

We systematically study the scaling properties of OPD in math reasoning.
OPD dynamics form a two-phase process: a clean useful-transfer phase, in which gold score ($G$) is approximately linear in the square root of token-level reverse KL ($d$), followed by a noisy tail.
Joint power laws in student scale, teacher scale, and teacher gold score predict peak gold score ($G_{\mathrm{peak}}$) and, more approximately, useful-transfer slope for both OPD variants.
Peak student error is nearly proportional to teacher remaining error, with smaller teachers transferring better at a matched score.
Bootstrapping does not improve on direct transfer from the smallest expert, OffPD underperforms OPD, and an off-policy cold start harms weak-to-strong transfer.
OPD outcomes can thus be estimated from student scale, teacher scale, and teacher score before training, so RL expertise can be trained once at small scale and transferred predictably across a model family.

\clearpage

\subsection*{Ethics statement}

Distillation can propagate a teacher's spurious task knowledge, hallucinations, biases, and unsafe patterns into a student.
Also, (bootstrapped) weak-to-strong distillation may amplify the downstream effects of these defects even when aggregate task accuracy initially improves.
Gold-score evaluation on a held-out test set provides partial safeguards, but it does not establish that the resulting model is reliable or safe beyond the evaluated setting.
Before deployment, distilled models should undergo broader evaluations of bias, safety, robustness, and distribution shift.

\bibliographystyle{assets/plainnat}
\bibliography{main}

\clearpage
\preto\section{\FloatBarrier}
\appendix

% Appendix TOC
\renewcommand \thepart{}
\renewcommand \partname{}

\part{Appendix}
\parttoc
\clearpage

\begin{table}[t]
\caption{Summary of notations.}
\centering
\footnotesize
\begin{tabularx}{\linewidth}{lX}
\toprule
Symbol & Meaning \\
\midrule
$x,\gD$ &
Prompt and training-prompt distribution. \\
$y,y_{<t},y_t,|y|$ &
Complete rollout, prefix before token $t$, token $t$, and rollout length. \\
$\policy,\reference$ &
Student policy parameterized by $\theta$ and its fixed SFT reference. \\
$\teacher,\teacherbase$ &
RL-trained teacher and its pre-RL reference checkpoint. \\
$r_t,r_t^\Delta$ &
Vanilla-OPD teacher-induced token reward and Delta-OPD teacher-shift reward. \\
$A_t^{\mathrm V},A_t^\Delta$ &
Advantages of Vanilla-OPD and Delta-OPD, equal to the token rewards under the
zero-discount update. \\
$\gL_{\mathrm V},\gJ_{\mathrm V},\gJ_\Delta,\gL_{\mathrm{Off}}$ &
Sequence-level Vanilla-OPD loss, token-level Vanilla-OPD objective, Delta-OPD
objective, and OffPD loss. \\
$G(\theta),G(d)$ &
Gold score as a function of parameters or of the KL coordinate. \\
$P$ &
Teacher-induced proxy score, logged as the token-mean reward on training
rollouts. \\
$\delta_t,k_3$ &
Token log-probability difference and primary token-mean KL estimator. \\
$d$ &
KL-indexed amount of optimization, $d=\sqrt{k_3}$. \\
$d_{\mathrm{transfer}}$ &
Last checkpoint compatible with the initial line before three sustained lower
deviations; right-censored when no departure is observed. \\
$d_{\mathrm{peak}}$ & Earliest observed value of $d$ attaining maximum gold score. \\
$N_S,N_T,\widetilde N_S,\widetilde N_T^{\mathrm{eff}}$ &
Student scale, teacher scale, the student scale normalized by one billion
parameters, and the teacher scale capped at student scale and likewise
normalized. \\
$G_T,G_T^{\mathrm{eff}}$ &
Held-out gold accuracy of the teacher RL endpoint, and of the teacher capped at
student scale. \\
$G_{\mathrm{peak}}$ & Peak gold score of a trajectory. \\
$A,\alpha,\beta,\zeta$ &
Joint peak-law coefficients (\cref{eq:teacher_score_scale}); $\zeta$ acts on
the capped-teacher error $1-G_T^{\mathrm{eff}}$. \\
$B,\gamma,\delta,\xi$ &
Joint transfer-rate-law coefficients (\cref{eq:teacher_score_scale}), fitted
independently for Vanilla-OPD and Delta-OPD. \\
$c,m$ & Intercept and slope of the local linear fit $G(d)=c+md$. \\
$G_{\mathrm V}(d),G_\Delta(d),\Delta G_{\mathrm V},\Delta G_\Delta$ &
Method-specific gold scores and gains over each run's own initialization. \\
$\gM_1,\dots,\gM_n$ &
Models along the bootstrapping chain in increasing scale; $\gM_1$ is the 0.5B RL
endpoint and each later $\gM_i$ is the OPD product taught by $\gM_{i-1}$. \\
\bottomrule
\end{tabularx}
\end{table}

%%%%%%%%%%%%%%%%%%%%%%%%%%%%%%%%%%%%%%%%%%%%%%%%%%%%%%%%%%%%%%%%%%%%%%%%%%%%%%%
%%%%%%%%%%%%%%%%%%%%%%%%%%%%%%%%%%%%%%%%%%%%%%%%%%%%%%%%%%%%%%%%%%%%%%%%%%%%%%%
\section{Discussion}
\label{sec:discussion}

\mypar{Capability dynamics are indexed by KL.}
Across the Qwen2.5 grid, every trajectory begins with useful transfer, while
later checkpoints exhibit attenuated improvement, saturation, or regression.
Both $d_{\mathrm{transfer}}$ and $d_{\mathrm{peak}}$ are properties of the
teacher--student pair.
Their reversals across scale show why a universal optimizer-step budget is
insufficient.

\mypar{Teacher quality is more than teacher size.}
The monotone peak trends for 7B and 14B students coexist with counterexamples at
smaller scales.  A useful scaling model should therefore condition on measured
teacher competence as well as parameter count.
Within the RL-endpoint grid the two are collinear at $r=-0.9999$ and cannot be
distinguished.
The bootstrapped chains break this collinearity with OPD-product teachers whose
scores sit off the size trend, and the resulting joint law
(\cref{eq:teacher_score_scale}) attributes peak error mainly to teacher
remaining error, with a negative size exponent at matched score
(\cref{tab:teacher_variable}).
A held-out comparison confirms this out of sample: an intermediate 3B
checkpoint scoring slightly above the 1.5B RL endpoint teaches the 7B
student worse, as only the joint law predicts (\cref{tab:intermediate_teacher}).
The bootstrapping comparison (\cref{sec:bootstrap}) shows the same dissociation
at the trajectory level: the 1.5B OPD teacher outscores the 0.5B RL expert yet
teaches the~3B~student~worse.

\mypar{Prediction connects the two regimes.}
The first 30 Vanilla-OPD and first 40 Delta-OPD observations estimate local transfer
rate.
Across scale, remaining peak error follows a stable joint power law in student
scale and teacher score, while transfer extent stays within a narrow band.
Peak capability therefore follows scale and teacher quality, whereas the
extent of the regular regime behaves as an approximately scale-free KL budget.

\mypar{Regression and overoptimization are not synonymous.}
Gold degradation can arise because the student over-imitates a teacher that
lacks some of its capabilities.
Teacher-induced implicit-reward overoptimization additionally requires
fixed-prompt evidence that $P$ continues to improve as $G$ falls.
Collapse denotes an abrupt optimization failure.
The diagnostics separate three testable mechanisms.

\mypar{Implications for OPD design.}
The peak law forecasts attainable capability from student scale and measured
teacher quality, while the rate law quantifies how quickly a method spends KL
to approach it.
The empirical transfer budget supplies a separate KL allowance, after which
gold measurements determine whether additional optimization yields attenuated
gain, saturation, or regression.
At the currently observed scales,
Delta-OPD converts token-mean KL into local gold gain more rapidly than
Vanilla-OPD in 15 of 17 shared cells and improves peak gain in 12.
Additional seeds are needed to determine whether this descriptive advantage is stable.

\mypar{Toward scaling beyond human supervision.}
\citet{yang2026scaling} outline how large reasoning models may continue to improve as human supervision recedes from the learning loop.
Our results bear on the supervision side of this program: a compact RL expert transfers capability to much larger students at rates and peaks predictable from scale, and the returns to teacher scaling taper off near the student's own scale.
Within OPD, supervision competence therefore need not grow with the policy it trains.
Feature-learning theory supports this picture: weak-to-strong training can succeed by eliciting knowledge already latent in the strong model~\citep{awano2026feature}.
If this regularity persists at frontier scale, weak-to-strong OPD is a candidate mechanism for eliciting capability beyond the supervisor's own level, with human-level supervision as the fixed weak teacher.

%%%%%%%%%%%%%%%%%%%%%%%%%%%%%%%%%%%%%%%%%%%%%%%%%%%%%%%%%%%%%%%%%%%%%%%%%%%%%%%
%%%%%%%%%%%%%%%%%%%%%%%%%%%%%%%%%%%%%%%%%%%%%%%%%%%%%%%%%%%%%%%%%%%%%%%%%%%%%%%
\section{Limitations}
\label{app:limitations}

\mypar{Scope.}
The evidence is limited to one pretrained family (Qwen2.5, 0.5B--14B), one
math reasoning mixture, one canonical run per teacher--student pair, responses
of at most 2K tokens, and intentionally stopped trajectories.
Evaluation records are aggregate rather than per prompt, so the present
figures have neither prompt-bootstrap nor optimization-seed intervals, and
cell-bootstrap intervals quantify variation across scale pairs but not across
training seeds.
Whether the fitted coefficients transfer to other families, tasks, or
post-training recipes is untested.

\mypar{Underspecified variables.}
Several recipe constants are held fixed and may matter.
Rollout length is capped at 2K tokens, although prior work shows that rollout
length affects OPD training stability~\citep{li2026rethinking}, and results
may also depend on the shared SFT recipe, the teacher RL checkpoint, the KL
estimator, validation frequency, and the same-step logging convention.
The token-mean coordinate weights tokens rather than responses; neither
archive contains full-sequence $k_3$, so estimator robustness remains
untested.
Within the RL-endpoint grid, teacher scale and teacher competence vary
together; the bootstrapped chains break this collinearity, but with only five
auxiliary Vanilla-OPD cells and three Delta-OPD cells, whose slopes come from
more sparsely logged evaluations.
One matched-accuracy comparison distills the 7B student from an intermediate
3B teacher checkpoint and confirms the joint law's ordering
(\cref{tab:intermediate_teacher}); repeating such controlled comparisons across the
grid remains future work.

\mypar{Mechanism and theory.}
The local expansion in \cref{app:local_geometry} explains why initial transfer
is linear in $d$, but only its leading exponent.
The extent of the linear window, the heterogeneous tails that follow it, and
the power-law forms of the fitted coefficients remain empirical regularities
without a mechanism.
The Delta-OPD endpoint law has only five observed departures, the
transfer-rate power law does not outperform the raw-affine baseline under
leave-one-scale-out RMSE, and the fixed 30- and 40-observation windows are exploratory.

Early termination and finite evaluation frequency can right-censor
$d_{\mathrm{transfer}}$ and place $d_{\mathrm{peak}}$ at the observation
boundary.
Multiple seeds, denser evaluation, and prompt-level resampling would quantify
both sources of uncertainty.

%%%%%%%%%%%%%%%%%%%%%%%%%%%%%%%%%%%%%%%%%%%%%%%%%%%%%%%%%%%%%%%%%%%%%%%%%%%%%%%
%%%%%%%%%%%%%%%%%%%%%%%%%%%%%%%%%%%%%%%%%%%%%%%%%%%%%%%%%%%%%%%%%%%%%%%%%%%%%%%
\section{Extended related work}
\label{app:related}

\mypar{OPD variants.}
A large number of OPD variants have emerged recently~\citep{song2026survey,awesome-on-policy-distillation}.
Generalized Knowledge Distillation trains on student-generated sequences and supports multiple divergences~\citep{agarwal2024onpolicy}.
ExOPD changes the relative reward and KL weights~\citep{yang2026learning}, while relaxed OPD interprets the log-likelihood ratio as a token reward and introduces clipping and sampling controls for instability~\citep{ko2026relaxed}.
W2S-OPD, Direct-OPD, and OPD$^2$ independently use contrasts between teacher checkpoints to isolate a learned policy shift~\citep{yu2026w2sopd,feng2026directopd,heo2026opd2}.
OPRD amplifies the component of the student's verifier-driven policy gradient along the teacher-shift direction, rescaling only verifier-supported updates so that teacher guidance accelerates rather than redirects the student's own optimization.
It reaches higher performance with fewer student updates than RL and distillation baselines in successor-generation transfer and multi-teacher consolidation~\citep{park2026oprd}.
OPD has also been combined with RL on verifiable rewards, coupling dense token-level distillation with outcome-level correctness signals, using either a separate teacher~\citep{lin2026policy,wang2026distilled} or the policy itself given hindsight access to verified solutions~\citep{cai2026h,yang2026self,li2026unifying}.
Empirical analyses further examine OPD training dynamics, failure modes, and recipes~\citep{li2026rethinking,fu2026revisiting}.
We use the contrast family of objectives as an experimental condition, which we call Delta-OPD (\cref{eq:delta}), and compare methods at matched KL rather than matched optimizer step.

\mypar{Proxy overoptimization and reward quality.}
The closest empirical precedent is the gold--proxy scaling study of \citet{gao2023scaling}, which separates RL and best-of-$n$ functional forms and relates their coefficients to reward-model scale, data, policy scale, and KL regularization.
Direct-alignment work finds related overoptimization laws without an explicit RL loop~\citep{rafailov2024scaling}, and reward hacking in large models has been surveyed systematically~\citep{wang2026reward}.
Complementary studies ask which reward-model properties yield useful optimization signals~\citep{razin2025what}, how preference-data scale changes RLHF~\citep{shen2025exploring}, and how inference-time search can exploit reward models~\citep{khalaf2025inference}.
These results motivate the proxy diagnostics used to interpret regressing OPD trajectories.

\mypar{Offline teacher hacking versus online distillation.}
\citet{tiapkin2025on} define teacher hacking in a controlled oracle--teacher--student hierarchy: repeated fixed teacher data can degrade ground-truth fit, while online generation and data diversity mitigate the effect.
This differs from our on-policy rollouts, but supplies an important mechanism-level control.
\citet{sriraman2026behavior} study a noisy-expert model and prove a separation between offline imitation and online interaction.
Their findings motivate OPD under imperfect supervision; they do not imply the empirical scale or shape of capability-transfer dynamics.

\mypar{Weak-to-strong generalization.}
\citet{burns2024weak} establish the empirical weak-to-strong problem and show that strong students can outperform weak supervisors.
In a tractable high-dimensional regression model, \citet{ildiz2024highdim} derive precise risk asymptotics and show that weak supervision can improve finite-data performance without improving the data-scaling exponent.
\citet{awano2026feature} analyze two-layer networks in the feature-learning regime and show that weak-to-strong fine-tuning elicits features latent in the strong model's pretrained knowledge while avoiding the forgetting induced by standard supervised fine-tuning.
\citet{yuan2026incentivizing} show that off-policy fine-tuning on chain-of-thought traces from much weaker reasoners recovers a large fraction of the gains of RL.
\citet{jin2026weak} use the weak model as a critic for scalable oversight, distilling revisions of the strong model's own outputs guided by the weak critiques.
We study elicitation through on-policy token-level scoring and find that peak gold score improves with teacher scale only up to roughly the student's own scale, captured by the effective-teacher cap in the peak law.

\mypar{Weakly supervised learning.}
Weak-to-strong learning can be viewed as an instance of weakly supervised learning, the long-studied setting where training signals are incomplete, inexact, or inaccurate~\citep{zhou2018brief}, including programmatic supervision that aggregates noisy labeling sources~\citep{ratner2017snorkel,bach2017learning}.
Learning from noisy labels is the closest classical problem~\citep{song2023learning}, but a weak teacher's mistakes concentrate on the instances it lacks competence for---the instance-dependent regime that is substantially harder than uniform label noise~\citep{frenay2014classification}---so standard noise-correction techniques do not directly apply.
The setting also differs from semi-supervised learning, where ground-truth labels are available for a subset of the training data~\citep{chapelle2006semi}: here no ground-truth reasoning supervision exists at the student's level, and gold labels serve only for held-out evaluation.
Within this family, weak-to-strong OPD is distinctive in that the weak signal supervises tokens of the student's own rollouts rather than static labels or demonstration trajectories, so supervision quality evolves with the student policy.

\mypar{Scaling laws for LLM training.}
Foundational neural scaling laws model pretraining loss as a power law of model size, data, and compute, enabling compute-allocation predictions~\citep{kaplan2020scaling}.
\citet{hoffmann2022training} systematically study compute-optimal scaling laws for LLM pretraining, finding that model size and data scale should be scaled equally.
Unified functional forms extend such fits across architectures, tasks, and resource axes~\citep{caballero2026unified}, and \citet{weng2026scaling} reviews how seemingly minor fitting choices can change extrapolated predictions.
Moving from pretraining to adaptation, \citet{zhang2024finetuning} find a multiplicative power-law relationship between fine-tuning performance and the amount of fine-tuning data together with model size, pretraining-data size, or the number of trainable parameters.
Their comparison of full-model and parameter-efficient fine-tuning further shows that the preferred method depends on the task and data regime.
\citet{busbridge2025distillation} fit compute-allocation laws for pretraining distillation, including cases with an existing teacher and cases where teacher training is charged to the budget.
\citet{lu2026strong} further find that small or undertrained teachers improve larger pretraining students while stronger teachers saturate or reverse the gains, an offline analogue of the effective-teacher cap in our peak law.
These results predict performance across resource and method configurations, rather than along a single KL-indexed optimization trajectory.

Post-training studies predict still different objects.
\citet{gao2023scaling} model gold reward along a KL-indexed optimization
trajectory and find optimization-specific functional forms whose coefficients
vary smoothly with reward-model scale.
\citet{rafailov2024scaling} extend this perspective to direct-alignment
objectives, where KL-indexed degradation can arise without a separately trained
proxy reward model.
By contrast, \citet{shen2026reasoning} connect the pretrained state to returns
from subsequent RL: pretraining loss predicts performance at fixed RL compute,
and early RL improvement rates vary systematically with pretraining data.

\citet{khatri2026art} instead fit sigmoidal performance--compute curves for LLM
RL and distinguish changes in asymptotic performance from changes in compute
efficiency across training recipes.
Their extrapolation from smaller runs makes RL performance predictable in a
compute coordinate, whereas our trajectories use token-mean KL as the amount
of optimization.
Collectively, these scaling properties concern different response variables
and optimization coordinates;
their functional forms should therefore not be transferred directly across
settings.

Our targets are how teacher scale, student scale, and the OPD objective shape the training dynamics: how much capability transfers, and how fast.
The local trajectory fits define these response variables; leave-one-scale-out
prediction determines whether their cross-pair relationships constitute
scaling laws.
This standard follows \citet{choshen2025hitchhiker}, who show that checkpoint
selection, model-size proximity, architecture, and seed variation materially
affect scaling estimates.
We therefore use leave-one-scale-out pair splits, checkpoint-level prediction,
simple baselines, architecture controls, and seed-aware uncertainty.
We report cross-objective results at matched KL rather than treating them as
matched-compute or algorithm-independent efficiency comparisons, since
objectives can accumulate KL at different rates for the same compute budget.

%%%%%%%%%%%%%%%%%%%%%%%%%%%%%%%%%%%%%%%%%%%%%%%%%%%%%%%%%%%%%%%%%%%%%%%%%%%%%%%
%%%%%%%%%%%%%%%%%%%%%%%%%%%%%%%%%%%%%%%%%%%%%%%%%%%%%%%%%%%%%%%%%%%%%%%%%%%%%%%
\section{Local geometry of capability transfer}
\label{app:local_geometry}

This section derives \cref{eq:local_geometry} for the population counterpart of
the token-mean estimator in \cref{eq:k3}.
The result explains the leading exponent for both objectives; it does not claim
that the empirical fitting windows are asymptotically small.

\subsection{Statement}
\label{app:local_geometry_statement}

For any differentiable initial path with update direction $h$ and nonzero
token-Fisher norm, local KL geometry~\citep{schulman2015trust} gives
\begin{equation}
 G(d)=G_0+md+O(d^2),\qquad
 m=\sqrt{2}\,\frac{g_0^\top h}
 {\sqrt{h^\top F_{\mathrm{tok}}h}},
 \label{eq:local_geometry}
\end{equation}
where $g_0=\nabla_\theta G(\theta_0)$ and $F_{\mathrm{tok}}$ is the token-weighted
Fisher matrix.
Gold score changes to first order in the perturbation while KL changes to
second order, so both objectives share the leading exponent and their slopes
differ through the Fisher-normalized alignment of $h$ with $g_0$.
How far linearity persists, and the ensuing $d_{\mathrm{transfer}}$ and
$d_{\mathrm{peak}}$, are empirical properties.

\subsection{A local square-root-KL law}

Let $\theta_0$ parameterize $\reference$, and define the reference score
$s_0(x,y_{<t},y_t)=
\nabla_\theta\log\policy(y_t\mid y_{<t},x)|_{\theta_0}$.
The population divergence corresponding to token-mean aggregation is
\begin{equation}
 D_{\mathrm{tok}}(\theta)
 =\frac{
 \E_{\substack{x\sim\gD,\,y\sim\policy(\cdot\mid x)}}
 \left[\sum_{t=1}^{|y|}
 \KL\!\left(\policy(\cdot\mid x,y_{<t})
 \Vert\reference(\cdot\mid x,y_{<t})\right)\right]}
 {\E_{\substack{x\sim\gD,\,y\sim\policy(\cdot\mid x)}}[|y|]}.
 \label{eq:population_token_kl}
\end{equation}
Indeed, for $\delta_t=\log\reference(y_t\mid x,y_{<t})-
\log\policy(y_t\mid x,y_{<t})$,
\begin{equation}
 \E_{y_t\sim\policy(\cdot\mid y_{<t},x)}[e^{\delta_t}-\delta_t-1]
 =\KL(\policy\Vert\reference),
 \label{eq:k3_population_identity}
\end{equation}
because $\E_{\policy}e^{\delta_t}=1$.
Thus \cref{eq:population_token_kl} is the population quantity estimated by the
logged token-mean $k_3$ statistic.

Consider a one-sided, differentiable training path,
\begin{equation}
 \theta(\tau)=\theta_0+\tau h+O(\tau^2),
 \qquad \tau\geq0.
 \label{eq:local_path}
\end{equation}
Define the token-weighted Fisher matrix
\begin{equation}
 F_{\mathrm{tok}}
 =\frac{\E_{\substack{x\sim\gD,\,y\sim\reference(\cdot\mid x)}}
 \left[\sum_{t=1}^{|y|}
 s_0(x,y_{<t},y_t)s_0(x,y_{<t},y_t)^\top\right]}
 {\E_{\substack{x\sim\gD,\,y\sim\reference(\cdot\mid x)}}[|y|]}.
 \label{eq:token_fisher}
\end{equation}
The conditional KL is zero at $\theta_0$, its first derivative is zero, and its
Hessian is the conditional Fisher matrix~\citep{schulman2015trust}.
Consequently,
\begin{equation}
 D_{\mathrm{tok}}(\theta(\tau))
 =\frac{\tau^2}{2}h^\top F_{\mathrm{tok}}h+O(\tau^3).
 \label{eq:token_kl_expansion}
\end{equation}
The on-policy prefix distribution also changes with $\theta$, but this drift
contributes only $O(\tau^3)$: its $O(\tau)$ change multiplies a conditional KL
that is already $O(\tau^2)$.

Let $G(\theta)$ be expected sampled mean@1 accuracy, rather than its finite
validation-set estimate, and write $g_0=\nabla_\theta G(\theta_0)$.
Smoothness gives
\begin{equation}
 G(\theta(\tau))=G_0+\tau g_0^\top h+O(\tau^2).
 \label{eq:gold_path_expansion}
\end{equation}
If $h^\top F_{\mathrm{tok}}h>0$, then
$d=\tau\sqrt{h^\top F_{\mathrm{tok}}h/2}+O(\tau^2)$.
Inverting this relation and substituting it into
\cref{eq:gold_path_expansion} proves \cref{eq:local_geometry}.
Positive initial transfer is equivalent to $g_0^\top h>0$.
If this inner product vanishes, the linear coefficient is zero and the leading
gold-score change can instead be quadratic in $d$.

\subsection{Method-specific directions and slopes}

At initialization, define the immediate token signals
\begin{equation}
 \begin{aligned}
 r_{\mathrm V}(x,y_{<t},y_t)
 &=\log\teacher(y_t\mid y_{<t},x)
 -\log\reference(y_t\mid y_{<t},x),\\
 r_\Delta(x,y_{<t},y_t)
 &=\log\teacher(y_t\mid y_{<t},x)
 -\log\teacherbase(y_t\mid y_{<t},x).
 \end{aligned}
 \label{eq:local_signals}
\end{equation}
Under the actor-update convention that sampled prefixes are held fixed within
an update, the population policy-gradient signal is
\begin{equation}
 b=\frac{\E_{\substack{x\sim\gD,\,y\sim\reference(\cdot\mid x)}}
 \left[\sum_{t=1}^{|y|}s_0(x,y_{<t},y_t)r(x,y_{<t},y_t)\right]}
 {\E_{\substack{x\sim\gD,\,y\sim\reference(\cdot\mid x)}}[|y|]},
 \qquad h=B_0b,
 \label{eq:method_tangent}
\end{equation}
where $B_0$ represents the optimizer's local preconditioning.
The reference-KL penalty has zero gradient at $\theta_0$, so it does not alter
this initial tangent.
Setting $r=r_{\mathrm V}$ or $r=r_\Delta$ selects the Vanilla-OPD or Delta-OPD
direction $h$ without changing the leading power of $d$.
By \cref{eq:local_geometry}, the local slope measures the Fisher-normalized
alignment of this direction with the gold gradient, rather than the magnitude
of the token reward.

\subsection{Assumptions and scope}

The derivation assumes common support and differentiable log probabilities; a
locally differentiable generated-prefix distribution with finite moments; a
differentiable initial optimizer path with nonzero Fisher norm; and a smooth
expected gold utility.
Relating the asymptotic result to the fitted windows additionally requires
rollout-policy lag, clipping, response-length changes, the Fisher geometry, and
the normalized gold alignment to vary slowly.
These assumptions are plausible for small early updates but are empirically
testable rather than guaranteed.
The theorem characterizes the leading term as $d\to0$.
Global shape, peak formation, estimator choice, and cross-method slope ordering
remain empirical properties.

%%%%%%%%%%%%%%%%%%%%%%%%%%%%%%%%%%%%%%%%%%%%%%%%%%%%%%%%%%%%%%%%%%%%%%%%%%%%%%%
%%%%%%%%%%%%%%%%%%%%%%%%%%%%%%%%%%%%%%%%%%%%%%%%%%%%%%%%%%%%%%%%%%%%%%%%%%%%%%%
\section{Hyperparameters}
\label{app:hyperparameters}

All SFT, RL, and OPD training runs are implemented on verl v0.8.0~\citep{sheng2025hybridflow}.

\subsection{SFT hyperparameters}

\begin{table}
\caption{Hyperparameters for Qwen2.5 SFT models.}
\label{tab:sft_hyperparameters}
\footnotesize
\centering
\setlength{\tabcolsep}{4pt}
\begin{tabular}{lccccc}
\toprule
Hyperparameter & Qwen2.5-0.5B & Qwen2.5-1.5B & Qwen2.5-3B & Qwen2.5-7B & Qwen2.5-14B \\
\midrule
Learning rate & 2.00e-5 & 1.52e-5 & 1.28e-5 & 1.03e-5 & 8.70e-6 \\
Batch size & \multicolumn{5}{c}{200} \\
Epochs & \multicolumn{5}{c}{2} \\
Samples per epoch & \multicolumn{5}{c}{300K} \\
Learning rate schedule & \multicolumn{5}{c}{Linear decay to 1/10 maximum LR} \\
Warmup ratio & \multicolumn{5}{c}{3\%} \\
Max response length & \multicolumn{5}{c}{2048} \\
Optimizer & \multicolumn{5}{c}{AdamW} \\
Weight decay & \multicolumn{5}{c}{0.1} \\
\bottomrule
\end{tabular}
\end{table}

For SFT, the models share the majority of hyperparameters except for the learning rate (\cref{tab:sft_hyperparameters}).
The learning rate scales as the inverse fourth root of parameter count ($N$):
\begin{equation}
\mathrm{LR}(N)=2\times10^{-5}
  \left(\frac{N}{0.5\text{B}}\right)^{-1/4}.
\end{equation}

\subsection{RL hyperparameters}

\begin{table}
\caption{Hyperparameters for Qwen2.5 RL teacher models.}
\label{tab:rl_hyperparameters}
\footnotesize
\centering
\begin{tabular}{lc}
\toprule
Hyperparameter & Value \\
\midrule
Learning rate & 1e-6 \\
Learning rate schedule & Constant \\
Global batch size & 256 \\
Global mini-batch size & 64 \\
GRPO group size & 8 \\
Epochs & 10 \\
Rollout temperature & 1.0 \\
Rollout top-p & 1.0 \\
Rollout top-k & -1 \\
Max rollout length & 2048 \\
KL penalty strength & 0.0 \\
Optimizer & AdamW \\
Weight decay & 0.01 \\
\bottomrule
\end{tabular}
\end{table}

All RL teachers share the hyperparameters in \cref{tab:rl_hyperparameters}.

\subsection{OPD hyperparameters}

\begin{table}
\caption{Hyperparameters for OPD on Qwen2.5 models.}
\label{tab:opd_hyperparameters}
\footnotesize
\centering
\begin{tabular}{lc}
\toprule
Hyperparameter & Value \\
\midrule
Learning rate & 1e-6 \\
Learning rate schedule & Constant \\
Batch size & 256 \\
Epochs & $\leq$10 \\
Max rollout length & 2048 \\
Optimizer & AdamW \\
Weight decay & 0.01 \\
\bottomrule
\end{tabular}
\end{table}

All OPD runs share the hyperparameters in \cref{tab:opd_hyperparameters}.
The intermediate-teacher validation run (\cref{app:scale_laws}) reuses this recipe for the 7B student and changes only the teacher checkpoint.

\subsection{Evaluation hyperparameters}

\begin{table}
\caption{Hyperparameters for evaluation.}
\label{tab:eval_hyperparameters}
\footnotesize
\centering
\begin{tabular}{lc}
\toprule
Hyperparameter & Value \\
\midrule
Temperature & 1.0 \\
Top-p & 1.0 \\
Top-k & -1 \\
Max rollout length & 2048 \\
\bottomrule
\end{tabular}
\end{table}

\Cref{tab:eval_hyperparameters} lists the sampling configuration shared by all gold-score evaluations.

\subsection{Random seeds}
\label{app:seeds}

Every SFT, RL, and OPD run in this study uses a single random seed, and we do not repeat any configuration across seeds.
The primary reason is compute.
The study spans five model scales with 25 Vanilla-OPD cells, 17 Delta-OPD cells, five RL teachers, and the bootstrapping and on-policy-supervision conditions, so each additional seed would multiply the training and evaluation cost of the entire grid.
Single runs per configuration are also standard in scaling-law studies, whose statistical support comes from regularity across many configurations rather than from per-configuration replication.
\citet{kaplan2020scaling} train one model per configuration and estimate seed-level loss variation at roughly 0.02 nats, small against their fitted trends; \citet{hoffmann2022training} fit compute-optimal laws on over 400 single-run models; and \citet{busbridge2025distillation} report one distillation run per teacher--student configuration.
Our preliminary trial reruns show OPD trajectories to be more stable than RL trajectories, and the cell-bootstrap intervals reported with the fitted laws quantify cross-pair variation rather than seed uncertainty (\cref{app:limitations}).

%%%%%%%%%%%%%%%%%%%%%%%%%%%%%%%%%%%%%%%%%%%%%%%%%%%%%%%%%%%%%%%%%%%%%%%%%%%%%%%
%%%%%%%%%%%%%%%%%%%%%%%%%%%%%%%%%%%%%%%%%%%%%%%%%%%%%%%%%%%%%%%%%%%%%%%%%%%%%%%
\section{Qwen2.5 math trajectory grid}
\label{app:curves}

\begin{figure}[t]
\centering
\includegraphics[width=\linewidth]{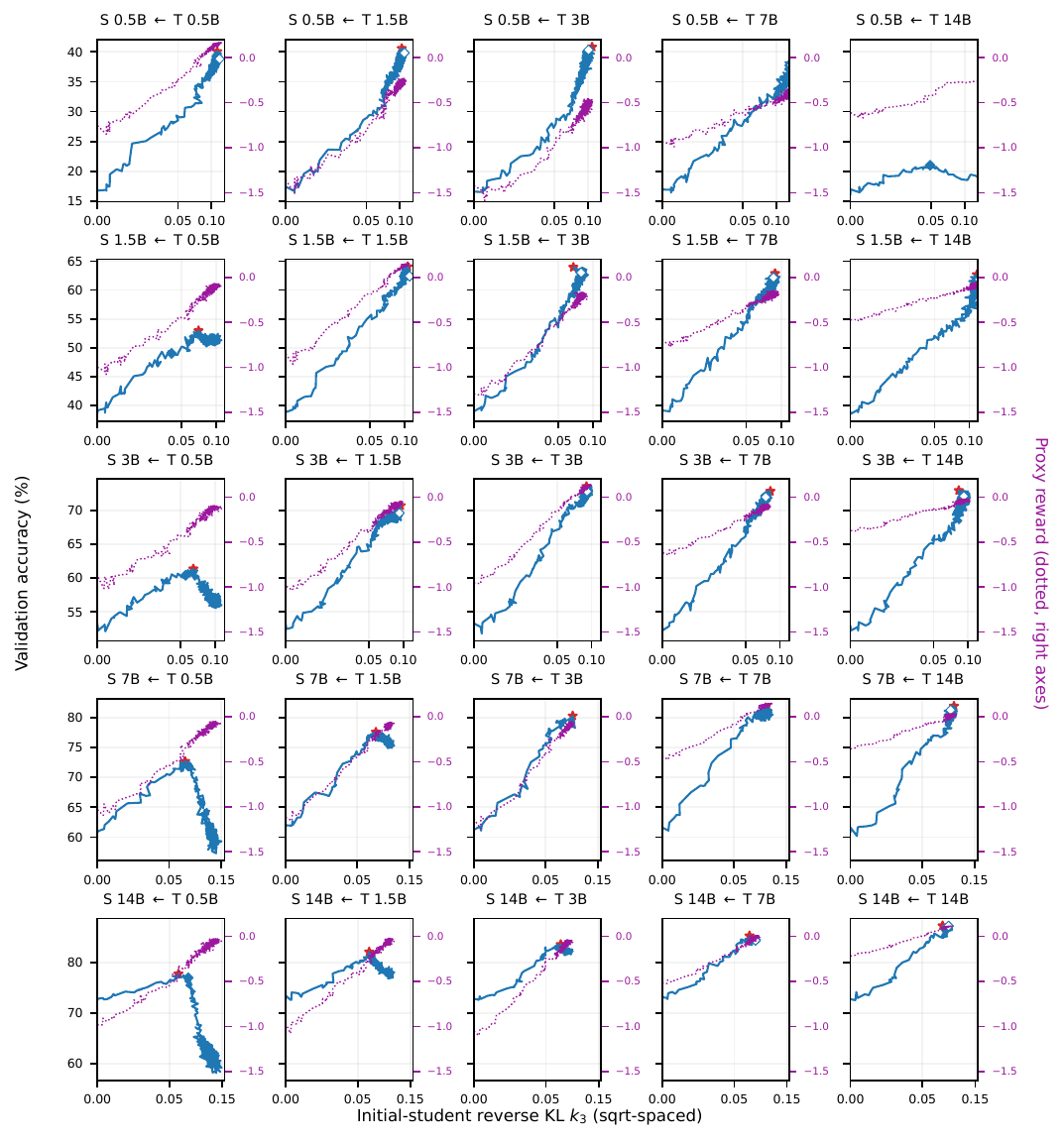}
\caption{Qwen2.5 mathematics trajectory grid with token-mean-KL ticks on
square-root-spaced axes.  Diamonds mark transfer endpoints (hollow when
right-censored), and stars mark global gold-score maxima.
Dotted purple curves show the teacher-induced proxy reward on right-hand axes
for the pairs where it was logged.
Rows share the student scale and columns share the teacher scale; panels
within a row share axis ranges.}
\label{fig:all_curves}
\end{figure}

\begin{figure}[t]
\centering
\includegraphics[width=\linewidth]{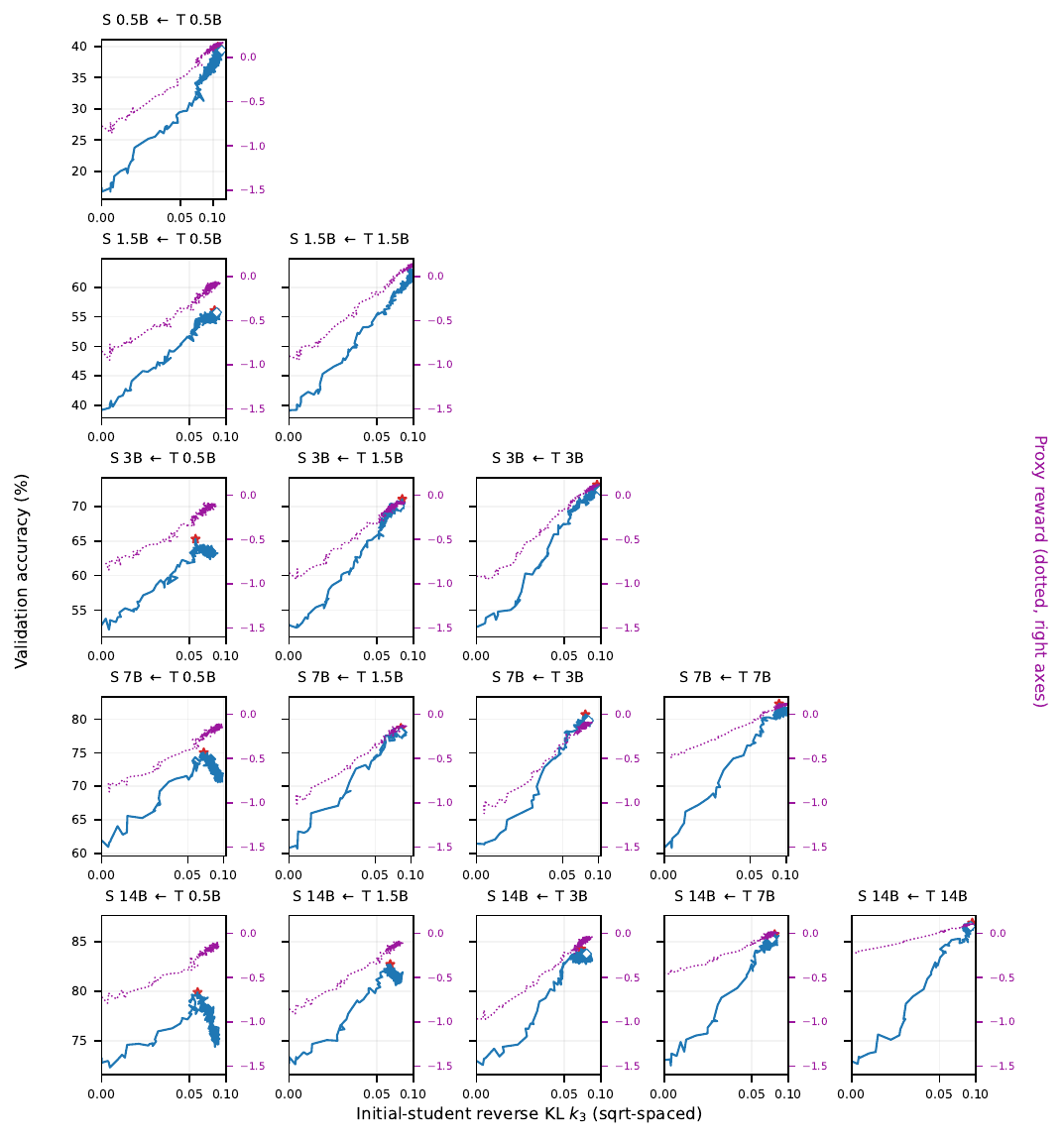}
\caption{Delta-OPD trajectory grid over weak-to-strong and same-base cells
with token-mean-KL ticks on square-root-spaced axes.
Diamonds mark transfer endpoints (hollow when right-censored), and stars mark
global gold-score maxima.
Dotted purple curves show the token-mean Delta-OPD reward on right-hand axes,
with the same ticks as \cref{fig:all_curves}.
Rows share the student scale and columns share the teacher scale; panels
within a row share axis ranges.}
\label{fig:delta_all_curves}
\end{figure}

\Cref{fig:all_curves} reports every canonical pair with the same alignment as
the main text.
Vanilla-OPD fit windows use the first 30 observations.
\Cref{fig:delta_all_curves} reports the corresponding grid for the 15
weak-to-strong and same-base Delta-OPD runs, whose fit window is 40
observations (\cref{sec:extensions}); the two strong-to-weak controls are omitted.

\FloatBarrier

%%%%%%%%%%%%%%%%%%%%%%%%%%%%%%%%%%%%%%%%%%%%%%%%%%%%%%%%%%%%%%%%%%%%%%%%%%%%%%%
%%%%%%%%%%%%%%%%%%%%%%%%%%%%%%%%%%%%%%%%%%%%%%%%%%%%%%%%%%%%%%%%%%%%%%%%%%%%%%%
\section{Expanded Delta-OPD and RL results}
\label{app:expanded_results}

\mypar{Complete Delta-OPD comparison.}
\Cref{tab:delta_full} gives the per-pair statistics summarized in the main
text.
The five observed transfer endpoints occur in four weak-to-strong cells and
one same-base cell; the remaining twelve endpoints are right-censored.

\begin{table}[!htbp]
\caption{Per-pair Delta-OPD results.
Peak gains are percentage points relative to each run's own initialization;
endpoint status is observed (Obs.) or right-censored (Cens.).}
\label{tab:delta_full}
\centering
\small
\setlength{\tabcolsep}{3.2pt}
\begin{tabular}{llrrrrrl}
\toprule
Pair & Relation & $m_{\mathrm V}$ & $m_\Delta$ & $R^2_\Delta$
& Gain V & Gain $\Delta$ & Endpoint \\
\midrule
0.5B$\leftarrow$0.5B & Same-base & .573 & .557 & .985 & 23.3 & 22.6 & Cens. \\
0.5B$\leftarrow$1.5B & S2W & .539 & .570 & .984 & 23.3 & 23.6 & Cens. \\
1.5B$\leftarrow$0.5B & W2S & .564 & .554 & .984 & 14.2 & 16.8 & Cens. \\
1.5B$\leftarrow$1.5B & Same-base & .721 & .744 & .981 & 24.6 & 24.6 & Cens. \\
1.5B$\leftarrow$3B & S2W & .629 & .723 & .973 & 24.8 & 24.6 & Cens. \\
3B$\leftarrow$0.5B & W2S & .378 & .444 & .963 & 8.2 & 12.4 & Obs. \\
3B$\leftarrow$1.5B & W2S & .555 & .642 & .973 & 17.6 & 18.2 & Cens. \\
3B$\leftarrow$3B & Same-base & .618 & .711 & .944 & 20.9 & 20.5 & Cens. \\
7B$\leftarrow$0.5B & W2S & .415 & .506 & .967 & 11.7 & 13.1 & Obs. \\
7B$\leftarrow$1.5B & W2S & .558 & .630 & .982 & 16.0 & 17.8 & Cens. \\
7B$\leftarrow$3B & W2S & .670 & .754 & .964 & 20.2 & 19.3 & Cens. \\
7B$\leftarrow$7B & Same-base & .687 & .712 & .987 & 20.3 & 21.5 & Obs. \\
14B$\leftarrow$0.5B & W2S & .175 & .268 & .916 & 5.1 & 6.7 & Obs. \\
14B$\leftarrow$1.5B & W2S & .327 & .387 & .961 & 9.3 & 9.9 & Obs. \\
14B$\leftarrow$3B & W2S & .417 & .470 & .942 & 10.6 & 11.4 & Cens. \\
14B$\leftarrow$7B & W2S & .424 & .470 & .971 & 12.1 & 12.6 & Cens. \\
14B$\leftarrow$14B & Same-base & .498 & .515 & .964 & 14.5 & 14.6 & Cens. \\
\bottomrule
\end{tabular}
\end{table}

\begin{figure}[!htbp]
\centering
\includegraphics[width=\linewidth]{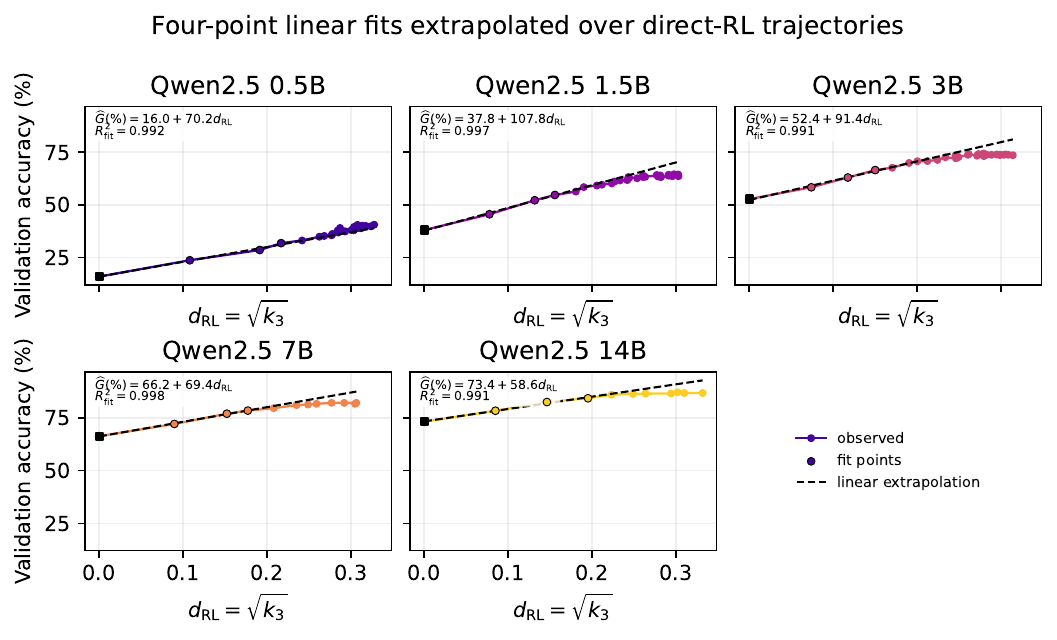}
\caption{Direct-RL gold score against token-mean reverse KL.
Each panel emphasizes the first four chronological observations, annotates the
free-intercept local fit, and extrapolates it across the remaining trajectory.
The black square marks the SFT initialization.}
\label{fig:rl_trajectories}
\end{figure}

\mypar{Direct-RL trajectories.}
\Cref{fig:rl_trajectories} shows the teacher-construction runs used for the
final-checkpoint reference in \cref{fig:scaling}.
Four-point local fits obtain $R^2\in[0.991,0.998]$, while extrapolation RMSE
over later checkpoints ranges from 0.016 to 0.041 accuracy units.
The growing error reflects diminishing marginal improvement along the direct-RL
trajectory.

\FloatBarrier

%%%%%%%%%%%%%%%%%%%%%%%%%%%%%%%%%%%%%%%%%%%%%%%%%%%%%%%%%%%%%%%%%%%%%%%%%%%%%%%
%%%%%%%%%%%%%%%%%%%%%%%%%%%%%%%%%%%%%%%%%%%%%%%%%%%%%%%%%%%%%%%%%%%%%%%%%%%%%%%
\section{Scaling law estimation and diagnostics}
\label{app:scale_laws}

\begin{figure}[!htbp]
\centering
\includegraphics[width=.78\linewidth]{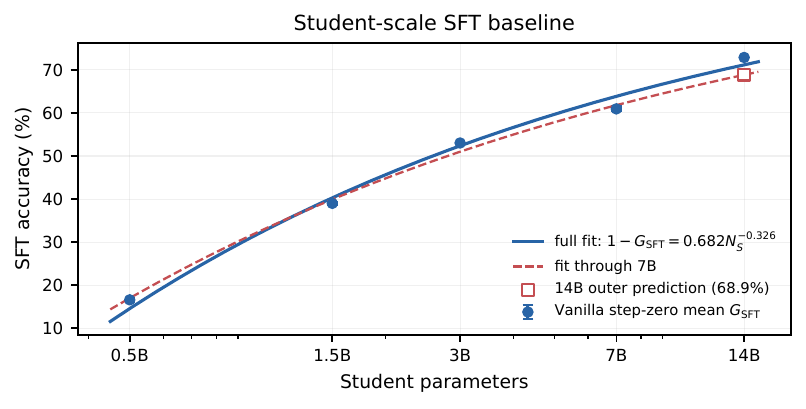}
\caption{Student-scale SFT baseline.
Points show equally weighted Vanilla step-zero means with within-scale standard
deviations.
The solid curve is the five-scale remaining-error fit in
\cref{eq:sft_scale}; the dashed curve is refit through 7B and its open square
is the 14B outer-scale prediction.}
\label{fig:sft_scaling}
\end{figure}

This appendix details the fitting protocols and diagnostics behind \cref{sec:scale_laws}.

\mypar{SFT baseline and peak capability.}
We average the Vanilla-OPD step-zero accuracies within each student scale to obtain
$G_{\mathrm{SFT}}(N_S)$ and fit the student-only remaining-error law
\begin{equation}
1-G_{\mathrm{SFT}}=0.682\,\widetilde N_S^{-0.326}.
\label{eq:sft_scale}
\end{equation}
Each of the five scale means averages five Vanilla-OPD cells.
Withholding the 14B mean gives a 4.0 percentage-point extrapolation error, the
refit law underpredicting the observed 14B baseline.
The vertical difference between this fitted baseline and a teacher-conditioned
peak in \cref{fig:scaling} visualizes capability added by OPD;
the fitted means and outer-scale check are shown in \cref{fig:sft_scaling}.
We use this difference for interpretation and retain per-run peak gains for the
matched-KL comparison, without fitting a separate gain law.

\begin{figure}[t]
\centering
\includegraphics[width=\linewidth]{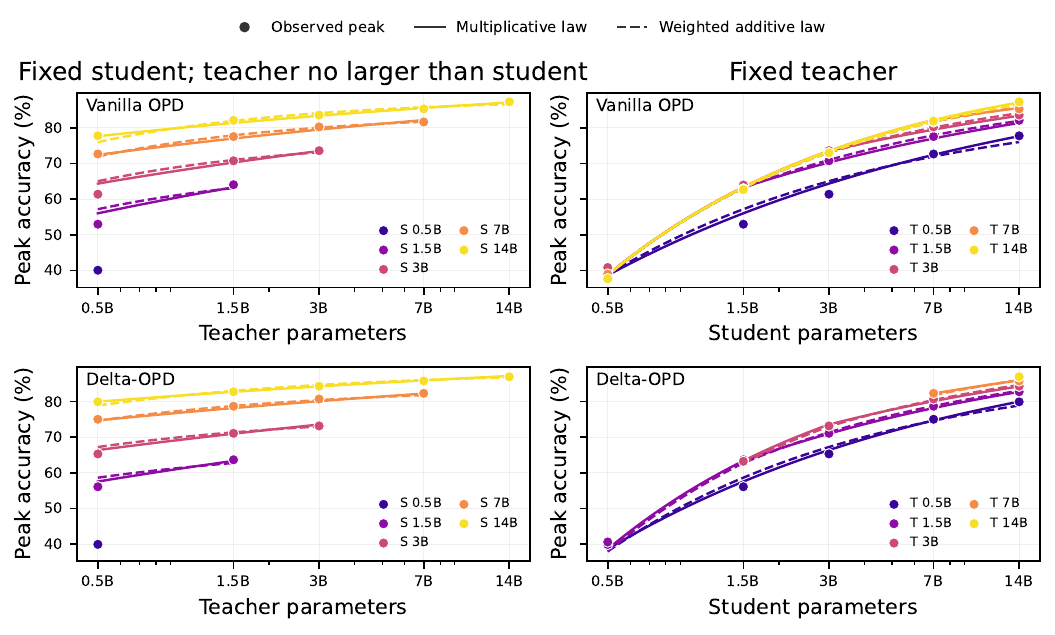}
\caption{Peak-performance scaling under the multiplicative and weighted-additive laws.
Rows separate Vanilla-OPD and Delta-OPD.
The left column fixes student scale and retains teachers no larger than the
student; the right column fixes teacher scale.
Points are observed peaks, solid curves are multiplicative-law fits, and dashed
curves are weighted-additive fits over each series' support.}
\label{fig:peak_scale_curves}
\end{figure}

\mypar{Model comparison.}
Peak-performance candidates use raw accuracy, logit accuracy, or log remaining
error with student-only, uncapped two-scale, and capped two-scale covariates.
Within the capped log-error family, we compare the multiplicative law in
\cref{eq:peak_scale}, a separately weighted additive law
$A_S\widetilde N_S^{-\alpha}+A_T(\widetilde N_T^{\mathrm{eff}})^{-\beta}$,
and the equal-amplitude sensitivity
$A[\widetilde N_S^{-\alpha}+(\widetilde N_T^{\mathrm{eff}})^{-\beta}]$.
The multiplicative law has the lowest mean leave-one-scale-out error and AICc for both
methods (\cref{tab:peak_model_comparison}).
\Cref{fig:peak_scale_curves} overlays the multiplicative and weighted-additive
fits on the observed peak series.

\begin{table}[!htbp]
\caption{Peak-error functional-form comparison.
S/T are leave-one-scale-out RMSE over student and teacher scales, in
percentage points.}
\label{tab:peak_model_comparison}
\centering
\small
\setlength{\tabcolsep}{5pt}
\begin{tabular}{llcc}
\toprule
Method & Peak-error law & S/T RMSE & AICc \\
\midrule
Vanilla & Multiplicative & 1.81 / 1.60 & $-155.6$ \\
& Weighted additive & 1.97 / 1.76 & $-140.8$ \\
& Equal additive & 4.02 / 2.67 & $-100.0$ \\
Delta & Multiplicative & .91 / .85 & $-108.3$ \\
& Weighted additive & 1.66 / 1.26 & $-88.7$ \\
& Equal additive & 3.99 / 3.04 & $-53.9$ \\
\bottomrule
\end{tabular}
\end{table}

\FloatBarrier

\begin{figure}[!t]
\centering
\includegraphics[width=\linewidth]{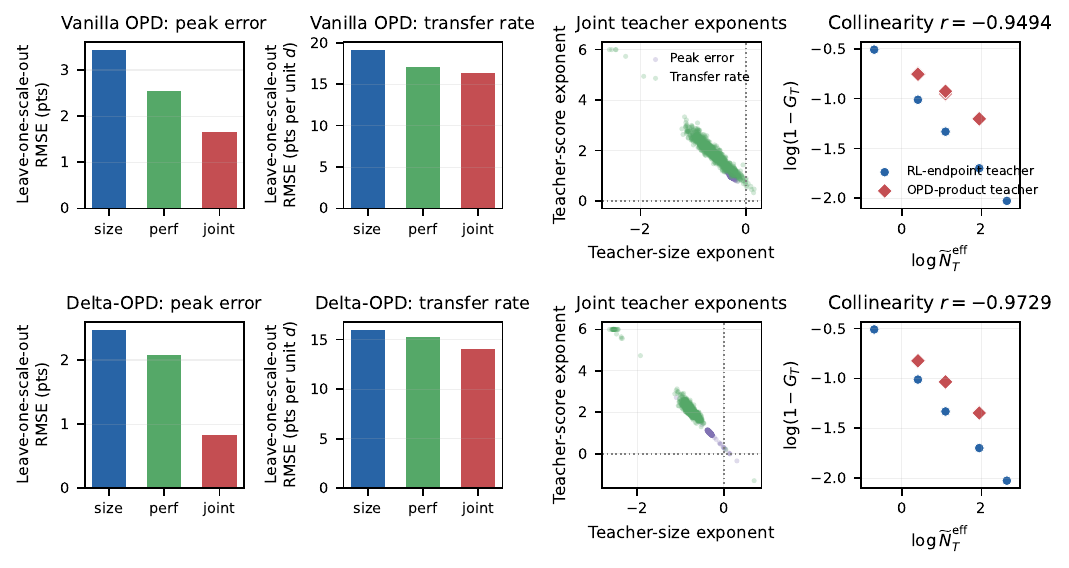}
\caption{Teacher size versus teacher performance, fitted on the grid plus the
auxiliary bootstrap-teacher cells; rows are Vanilla-OPD (top) and Delta-OPD
(bottom).
Left two columns: leave-one-scale-out RMSE of the size, performance, and
joint laws for the peak and rate targets.
Third column: bootstrap joint-law teacher exponents for both targets,
concentrated away from zero on anticorrelated ridges.
Right: capped teacher size against capped-teacher error; red diamonds mark
OPD-product teachers, which sit off the RL-endpoint size trend and break
the collinearity.}
\label{fig:teacher_variable}
\end{figure}

\mypar{Teacher size versus teacher performance.}
Parameter count is one of two natural teacher covariates; the teacher's own
held-out gold accuracy $G_T$ is the other.
For a fair single-covariate comparison we replace
$\widetilde N_T^{\mathrm{eff}}$ in \cref{eq:peak_scale} by the error of the
same capped teacher, and fit
\begin{equation}
1-G_{\mathrm{peak}}
=A\widetilde N_S^{-\alpha}
(1-G_T^{\mathrm{eff}})^{\zeta}.
\label{eq:perf_peak_scale}
\end{equation}
Within the RL-endpoint grid alone the two teacher covariates are collinear at
$r=-0.9999$, so the size and performance laws cannot be separated there.
The teacher-variable fits therefore add the bootstrap-chain cells of
\cref{sec:bootstrap}, whose teachers are OPD products with measured gold scores
below the size trend: five auxiliary Vanilla-OPD cells and three Delta-OPD cells.
Each chain stage also yields an initial slope, fitted on its method's standard
initial window; chain evaluations are logged more sparsely than the grid's, so these
windows span a longer $d$-range while remaining inside the useful-transfer
regime.
The added teachers reduce the collinearity to $r=-0.949$ for Vanilla-OPD and
$r=-0.973$ for Delta-OPD (\cref{fig:teacher_variable}, right).
For the peak target the joint law in \cref{eq:teacher_score_scale} becomes
identifiable for Vanilla-OPD, $\zeta=0.952$ with 95\% interval $[0.90,1.04]$
and $\beta=-0.267$ with interval $[-0.31,-0.23]$, and its leave-one-scale-out
RMSE improves to 1.66 points from 2.55 (score only) and 3.43 (size only)
(\cref{tab:teacher_variable}).
Peak student error is thus nearly proportional to the capped teacher's
remaining error, and at matched score the smaller teacher transfers better.
The three Delta-OPD chain cells identify the peak law as well: $\zeta=1.009$
with interval $[0.90,1.11]$ and $\beta=-0.325$ with interval $[-0.37,-0.28]$,
with leave-one-scale-out RMSE of 0.82 points from 2.07 (score only) and 2.47
(size only).
The rate target repeats the pattern with larger score exponents and noisier
fits: $\xi=1.90$ with interval $[0.93,2.84]$ and $\delta=-0.61$ with interval
$[-1.10,-0.10]$ for Vanilla-OPD, and $\xi=2.01$ with interval $[1.62,5.78]$
and $\delta=-0.73$ with interval $[-2.44,-0.54]$ for Delta-OPD, so a worse
teacher slows transfer much more strongly than it limits the~eventual~peak.

\begin{table}[t]
\caption{Teacher-size, teacher-performance, and joint laws for the peak and
rate targets, fitted on the grid cells plus the auxiliary bootstrap-teacher
cells (five for Vanilla-OPD, three for Delta-OPD).
Leave-one-scale-out RMSE is the mean of the student-scale and teacher-scale
hold-out splits, in accuracy points for the peak target and in points per unit
$d$ for the rate target.
The teacher exponent acts on $\widetilde N_T^{\mathrm{eff}}$ for the size law
and on $1-G_T^{\mathrm{eff}}$ for the performance law; the joint law fits
both, with 95\% bootstrap intervals in \cref{tab:scale_coefficients}.}
\label{tab:teacher_variable}
\footnotesize
\centering
\setlength{\tabcolsep}{2.5pt}
\begin{tabular}{lllccccc}
\toprule
Method & Target & Teacher law & Student exp. & Teacher exponents & Hold-out RMSE & AICc & $R^2$ \\
\midrule
\multirow{6}{*}{Vanilla-OPD}
 & \multirow{3}{*}{Peak} & Size & 0.278 & $\beta=0.147$ & 3.43 & $-101.2$ & 0.915 \\
 & & Perf. & 0.269 & $\zeta=0.425$ & 2.55 & $-129.7$ & 0.967 \\
 & & Joint & 0.304 & $\beta=-0.267,\ \zeta=0.952$ & 1.66 & $-190.2$ & 0.996 \\
\cmidrule{2-8}
 & \multirow{3}{*}{Rate} & Size & 0.240 & $\delta=0.217$ & 19.1 & $-13.0$ & 0.243 \\
 & & Perf. & 0.269 & $\xi=0.697$ & 17.0 & $-21.3$ & 0.427 \\
 & & Joint & 0.189 & $\delta=-0.609,\ \xi=1.896$ & 16.3 & $-28.2$ & 0.587 \\
\midrule
\multirow{6}{*}{Delta-OPD}
 & \multirow{3}{*}{Peak} & Size & 0.332 & $\beta=0.105$ & 2.47 & $-69.3$ & 0.929 \\
 & & Perf. & 0.320 & $\zeta=0.303$ & 2.07 & $-80.3$ & 0.961 \\
 & & Joint & 0.336 & $\beta=-0.325,\ \zeta=1.009$ & 0.82 & $-129.3$ & 0.998 \\
\cmidrule{2-8}
 & \multirow{3}{*}{Rate} & Size & 0.216 & $\delta=0.135$ & 16.0 & $-14.7$ & 0.354 \\
 & & Perf. & 0.242 & $\xi=0.438$ & 15.2 & $-19.5$ & 0.492 \\
 & & Joint & 0.203 & $\delta=-0.725,\ \xi=2.010$ & 14.1 & $-30.6$ & 0.757 \\
\bottomrule
\end{tabular}
\end{table}

\FloatBarrier

\mypar{Validation with an intermediate teacher checkpoint.}
The joint peak law predicts that when two teachers reach the same gold score, the larger one yields the worse student.
We test this prediction out of sample by distilling the 7B student from an intermediate checkpoint of the 3B teacher's RL run, taken at step 58, the earliest saved epoch, whose directly evaluated gold score of 66.0 approximately matches the 1.5B RL endpoint's 63.6.
The run reuses the canonical Vanilla-OPD recipe (\cref{app:hyperparameters}) and changes only the teacher checkpoint, and its cell enters no fit.
\Cref{tab:intermediate_teacher} compares observed peaks with the full-fit predictions of the three Vanilla-OPD peak laws, and \cref{fig:intermediate_teacher} overlays the trajectories (both in \cref{sec:scale_laws}).
The archived trajectory reaches update 430 of 580 and peaks at update 46, far before the truncation.

\begin{figure}[t]
\centering
\includegraphics[width=\linewidth]{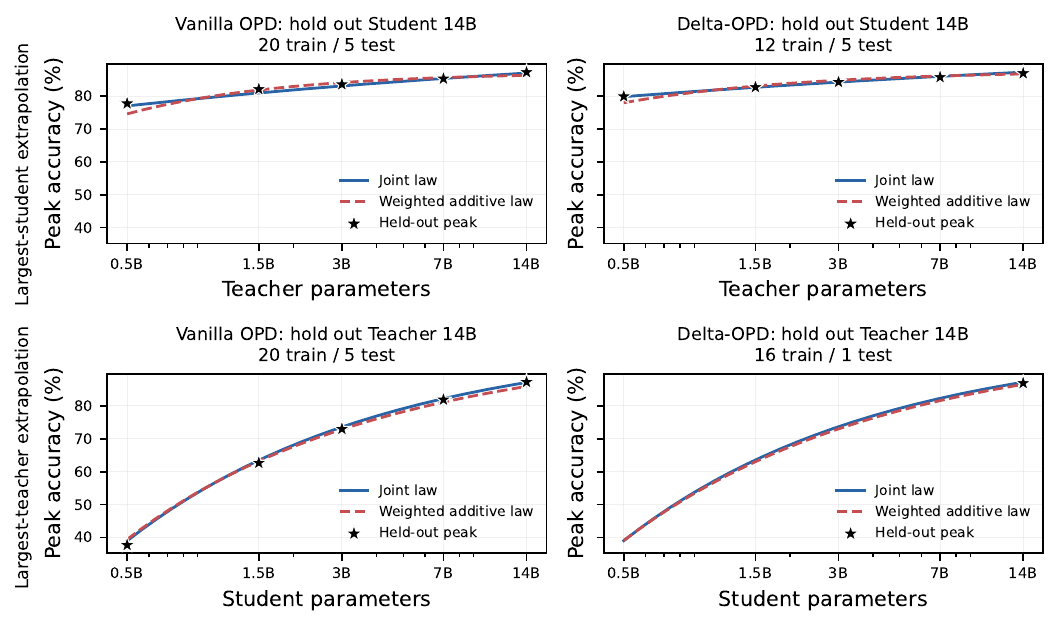}
\caption{Peak-law extrapolation after fitting only smaller scales.
Rows hold out the largest student or teacher scale; columns separate Vanilla-OPD
and Delta-OPD.
Stars are held-out peaks, solid curves are joint-law predictions, and dashed
curves are weighted-additive predictions.}
\label{fig:outer_scale_extrapolation}
\end{figure}

\mypar{Outer-scale extrapolation.}
We refit the size-only peak laws and the joint teacher law after withholding
the largest student scale and the largest observed teacher scale as separate
tests.
The largest-teacher scale is 14B for both methods.
All models share each split's grid test cells; the joint teacher law
additionally trains on the chain cells below the held-out scale.
The size-only joint law improves every axis--method comparison over the
additive law, and the joint teacher law extrapolates comparably, improving
the Vanilla-OPD holdouts while staying within 0.03 points of the Delta-OPD
ones (\cref{tab:outer_scale_extrapolation,fig:outer_scale_extrapolation}).
Bootstrap intervals resample training cells and measure fit stability rather
than variation across training seeds.
Nine replicates are non-identifiable and excluded; the other 11,991
model--split replicates converge and pass the rank check.

\begin{table}[t]
\caption{Outer-scale peak extrapolation.
Entries are accuracy-point RMSE with 95\% training-cell-bootstrap intervals.
Train counts refer to the size-only laws; the joint teacher law adds the
chain cells below the held-out scale (four Vanilla-OPD and two Delta-OPD for
the student holdout, five and three for the teacher holdout).}
\label{tab:outer_scale_extrapolation}
\small
\setlength{\tabcolsep}{4pt}
\centering
\begin{tabular}{llcccc}
\toprule
Holdout & Method & Train/Test & Joint & Weighted additive & Joint teacher \\
\midrule
Largest student & Vanilla & 20/5 & $.64\ [.33,2.27]$ & $1.50\ [.67,3.60]$ & $.55\ [.28,1.80]$ \\
& Delta & 12/5 & $.21\ [.12,.89]$ & $.92\ [.46,2.35]$ & $.20\ [.11,.96]$ \\
Largest teacher & Vanilla & 20/5 & $.75\ [.32,1.36]$ & $1.06\ [.67,1.64]$ & $.68\ [.33,1.25]$ \\
& Delta & 16/1 & $.29\ [.04,.72]$ & $.37\ [.02,1.49]$ & $.32\ [.07,.68]$ \\
\bottomrule
\end{tabular}
\end{table}

\FloatBarrier

\mypar{Direct accuracy-gain sensitivity.}
Direct, weighted-additive, and finite-horizon direct-RL-gap parameterizations of
peak gain produced less consistent outer-scale extrapolation than the separate
SFT and peak endpoints.
We therefore retain observed per-run gains for descriptive matched-KL
comparisons, but report no standalone gain law.
The exploratory fits and derived predictions remain available in the analysis
outputs for reproducibility.

\FloatBarrier

\begin{table}[t]
\caption{Scale-law validation under the joint teacher laws of
\cref{eq:teacher_score_scale}: peak uses outer-scale holdout errors (S/T) in
points; rate uses leave-one-scale-out RMSE (S/T) in accuracy per unit $d$.}
\label{tab:scale_validation}
\centering
\small
\setlength{\tabcolsep}{4pt}
\begin{tabular}{lccc}
\toprule
Method & Peak S/T & Rate S/T & Endpoint model \\
\midrule
Vanilla & .55 / .68 & .203 / .124
& two-scale $C=.37$ \\
Delta & .20 / .32 & .177 / .104
& two-scale $C=.34$ \\
\bottomrule
\end{tabular}
\end{table}

\Cref{tab:scale_validation} aggregates the validation summary of the joint
teacher laws across the peak, rate, and extent targets.
For transfer rate, the scale-only positive power law is also compared with a
raw-affine model using identical covariates and held-out cells.
The raw baseline retains a modest predictive advantage (leave-one-scale-out
RMSE $.186/.117$ against $.208/.126$ for Vanilla-OPD and $.181/.112$ against
$.196/.116$ for Delta-OPD), so the fitted exponents are read as an
interpretable scale summary rather than the best predictor.

\begin{figure}[t]
\centering
\includegraphics[width=.3\linewidth]{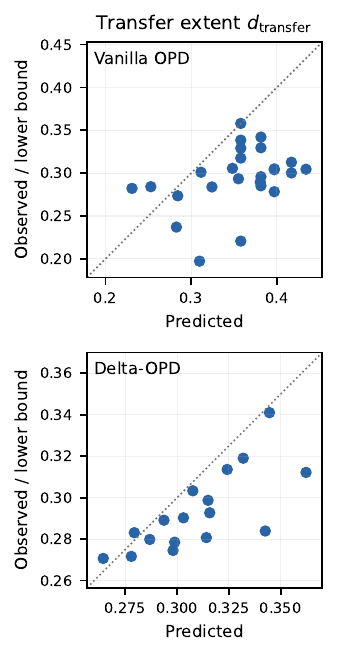}
\caption{Transfer-extent diagnostics.
Observed departures or censored lower bounds against the
accelerated-failure-time fits of \cref{eq:endpoint_fits} for
Vanilla-OPD (top) and Delta-OPD (bottom).}
\label{fig:extent_diagnostics}
\end{figure}

Endpoint candidates use log-normal accelerated-failure-time likelihoods with
constant, student-only, and two-scale predictors, fitted to the observed and
right-censored departures.
Mean leave-one-scale-out log likelihood over student and teacher splits
selects the capped two-scale
model for Vanilla-OPD, whose exponents are bootstrap-identifiable with 997 of
1,000 stable samples, and the uncapped two-scale model for Delta-OPD, whose
smaller exponents are not.
The Delta-OPD fit remains exploratory: five events yield 775 stable
cell-bootstrap samples out of 1,000.
The selected laws are
\begin{equation}
\operatorname{median}(d_{\mathrm{transfer},\mathrm{V}})
=0.37\,\widetilde N_S^{-0.13}
(\widetilde N_T^{\mathrm{eff}})^{0.19},
\qquad
\operatorname{median}(d_{\mathrm{transfer},\Delta})
=0.34\,\widetilde N_S^{-0.08}
\widetilde N_T^{\,0.05}.
\label{eq:endpoint_fits}
\end{equation}
Both laws vary the median budget by less than a factor of two across the
grid, shrinking slowly with student scale and growing slowly with teacher
scale, and \cref{fig:extent_diagnostics} plots their predictions against
observed or censored endpoints.
Given these small and partly unidentifiable exponents, the main body reports
the transfer extent as an approximately scale-free KL budget rather than a
fitted target.

\begin{table}[t]
\caption{Power-law coefficients with 95\% cell-bootstrap intervals.
Peak and rate report the joint teacher laws of \cref{eq:teacher_score_scale};
intervals quantify cross-pair variation rather than seed uncertainty.
Extent rows report the median-budget laws of \cref{eq:endpoint_fits},
$\operatorname{median}(d_{\mathrm{transfer}})=C\widetilde N_S^{-u}\widetilde N_T^{\,v}$,
with the teacher scale capped or uncapped as marked.}
\label{tab:scale_coefficients}
\centering
\footnotesize
\setlength{\tabcolsep}{3.5pt}
\begin{tabular}{lll}
\toprule
Method & Target & Coefficients \\
\midrule
Vanilla & Peak error &
$A=.97\,[.93,1.04],\ \alpha=.304\,[.286,.317],$ \\
& & $\beta=-.267\,[-.311,-.229],\ \zeta=.952\,[.898,1.035]$ \\
Delta & Peak error &
$A=1.02\,[.93,1.11],\ \alpha=.336\,[.327,.351],$ \\
& & $\beta=-.325\,[-.372,-.276],\ \zeta=1.009\,[.904,1.114]$ \\
SFT & Baseline &
$A_{\mathrm{SFT}}=.682\,[.680,.683],\ \alpha_{\mathrm{SFT}}=.326\,[.324,.328]$ \\
Vanilla & Rate &
$B=.116\,[.050,.263],\ \gamma=.189\,[.011,.306],$ \\
& & $\delta=-.61\,[-1.10,-.10],\ \xi=1.90\,[.93,2.84]$ \\
Delta & Rate &
$B=.129\,[.006,.181],\ \gamma=.203\,[.087,.304],$ \\
& & $\delta=-.73\,[-2.44,-.54],\ \xi=2.01\,[1.62,5.78]$ \\
Vanilla & Extent &
$C=.373\,[.305,.480],\ u=.132\,[.030,.328],\ v=.190\,[.036,.601]$ (capped) \\
Delta & Extent &
$C=.337\,[.309,.455],\ u=.080\,[.039,.106],\ v=.045\,[.018,.527]$ (uncapped) \\
\bottomrule
\end{tabular}
\end{table}

\mypar{Sensitivity.}
Replacing $\widetilde N_T^{\mathrm{eff}}$ by uncapped teacher size or omitting
teacher scale worsens leave-one-scale-out peak prediction.
Adding a learnable irreducible error to \cref{eq:peak_scale} places that term at
its zero lower bound for both methods, providing no evidence for an additional
floor over the observed scale range.
Endpoint-rule sensitivity remains governed by the confidence-band and
persistence grid in \cref{tab:endpoint_sensitivity}.

\FloatBarrier

%%%%%%%%%%%%%%%%%%%%%%%%%%%%%%%%%%%%%%%%%%%%%%%%%%%%%%%%%%%%%%%%%%%%%%%%%%%%%%%
%%%%%%%%%%%%%%%%%%%%%%%%%%%%%%%%%%%%%%%%%%%%%%%%%%%%%%%%%%%%%%%%%%%%%%%%%%%%%%%
\section{Endpoint and estimator sensitivity}
\label{app:statistics}

\mypar{Transfer-endpoint sensitivity.}
The primary endpoint uses a one-sided 95\% predictive band and three
consecutive lower deviations after the prespecified fit window.
The machine-readable analysis records the endpoint checkpoint, KL coordinate,
deviation start, and right-censoring status for every run.
Confidence levels of 90\%, 95\%, and 99\% crossed with persistence requirements
of two, three, and four checkpoints produce the sensitivity grid in
\cref{tab:endpoint_sensitivity}.

\begin{table}[t]
\centering
\footnotesize
\setlength{\tabcolsep}{5pt}
\caption{Endpoint-rule sensitivity: number of Vanilla-OPD trajectories with
an observed departure from the initial line.}
\label{tab:endpoint_sensitivity}
\begin{tabular}{lccc}
\toprule
One-sided band & Two deviations & Three deviations & Four deviations \\
\midrule
90\% & 12 & 9 & 9 \\
95\% & 9 & 9 & 9 \\
99\% & 8 & 8 & 8 \\
\bottomrule
\end{tabular}
\end{table}
Alternate alignment, training seeds, and prompt bootstrap resamples provide
additional validation axes.
Held-out endpoint evaluation reports point error for observed departures and
one-sided consistency for right-censored trajectories.

For the 17 Delta-OPD first-40 windows, no unconstrained quadratic has both
statistically identifiable negative curvature and a meaningful in-support
vertex under the prespecified condition-number and signal-to-noise checks.
Their later checkpoints are therefore presented as observed trajectories.

\mypar{Estimator sensitivity.}
The primary token-mean coordinate is defined in \cref{eq:k3}.
As a length-sensitive diagnostic, the archive also records the response mean
of tokenwise $k_3$ sums,
\begin{equation}
 k_{3,\mathrm{sum}}
 =\frac1n\sum_{i=1}^n\sum_t
 \left[\exp(\delta_{i,t})-\delta_{i,t}-1\right].
 \label{eq:k3-token}
\end{equation}
Here, $\delta_{i,t}$ is the token log-probability difference already defined
in \cref{eq:k3}.
This \texttt{actor/response\_reverse\_kl} scalar is not sequence-level KL:
full-sequence $k_3$ would exponentiate
$\delta_i=\sum_t\delta_{i,t}$ only after summation, and neither archive logs
that quantity.

The sensitivity analysis compares token-mean and token-summed coordinates
through correlations, curve overlays, endpoint changes, and local-fit
coefficients.
Full-sequence measurements remain a separate validation target because the
archives lack that scalar.

Future estimator validation will compare full-sequence $k_3$ with sampled
$k_1$ and partial full-vocabulary KL on common prompts and checkpoints.
Abrupt failures will be diagnosed using update KL, gradient norm, entropy, and
response length as a separate collapse category.

%%%%%%%%%%%%%%%%%%%%%%%%%%%%%%%%%%%%%%%%%%%%%%%%%%%%%%%%%%%%%%%%%%%%%%%%%%%%%%%
%%%%%%%%%%%%%%%%%%%%%%%%%%%%%%%%%%%%%%%%%%%%%%%%%%%%%%%%%%%%%%%%%%%%%%%%%%%%%%%
\section{Extended results for design choices}
\label{app:extensions}

\begin{figure}[!htbp]
\centering
\includegraphics[width=\linewidth]{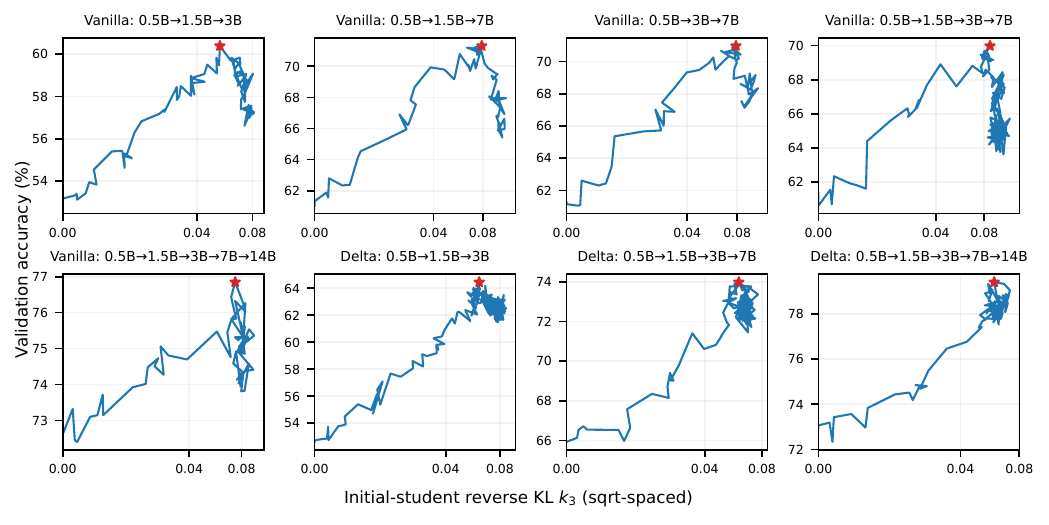}
\caption{Gold-score dynamics of the five Vanilla-OPD and three Delta-OPD
bootstrap chains.
Stars mark the observed peak.}
\label{fig:bootstrap_dynamics}
\end{figure}

\mypar{Bootstrap training dynamics.}
\Cref{fig:bootstrap_dynamics} shows the full gold-score trajectory of every bootstrapped run in \cref{sec:bootstrap} against its own KL-indexed training progress.
Most chains reach their peak early and then decline or saturate, mirroring the weak-to-strong dynamics of the direct grid.

\begin{figure}[t]
\centering
\includegraphics[width=\linewidth]{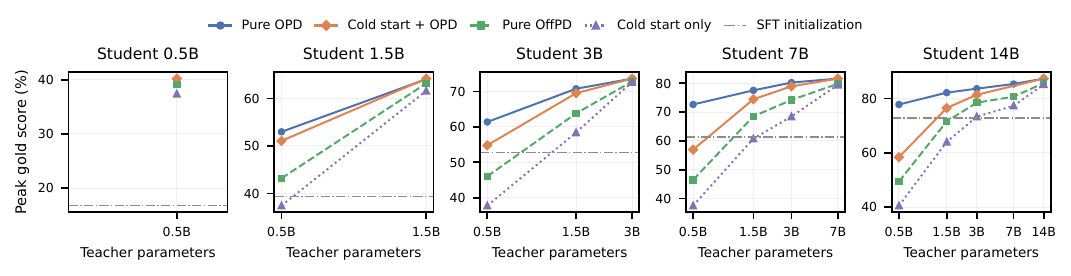}
\caption{Peak gold score under three degrees of on-policy supervision, per
student, over weak-to-strong and same-base cells.
Dash-dotted levels mark SFT initializations.}
\label{fig:onpolicyness_grid}
\end{figure}

\mypar{Full on-policyness grid.}
\Cref{fig:onpolicyness_grid} expands \cref{fig:onpolicyness} to every weak-to-strong and same-base cell, one panel per student.
The orderings reported in \cref{sec:onpolicyness} hold across the grid.

\begin{figure}[t]
\centering
\includegraphics[width=\linewidth]{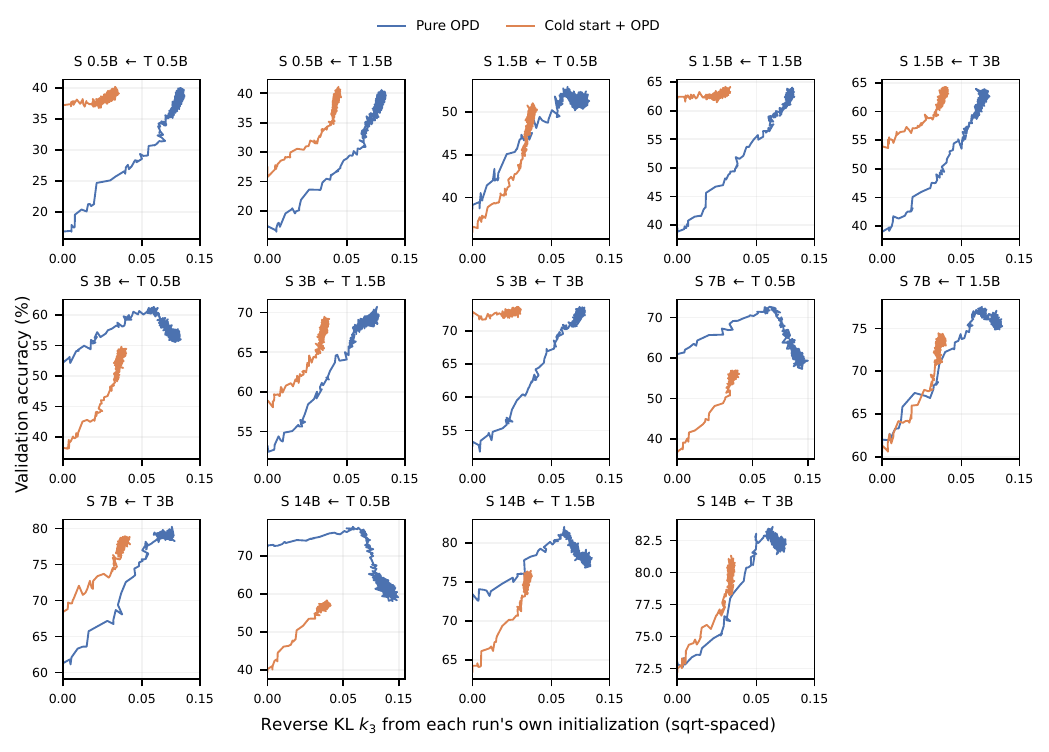}
\caption{Cold-start OPD against pure OPD on the 14 shared cells.
Each trajectory uses the reverse KL from its own initialization, so the two
curves in a panel have different reference policies.}
\label{fig:cold_start_dynamics}
\end{figure}

\mypar{Cold-start dynamics.}
\Cref{fig:cold_start_dynamics} overlays OPD from the off-policy cold start with pure OPD on the shared cells of \cref{sec:onpolicyness}, each in the KL coordinate measured from its own initialization.
In weak-to-strong cells the cold-started runs begin from a degraded initialization and track a lower trajectory throughout, so the peak deficit originates in the cold start rather than in the~subsequent~\mbox{on-policy}~phase.

%%%%%%%%%%%%%%%%%%%%%%%%%%%%%%%%%%%%%%%%%%%%%%%%%%%%%%%%%%%%%%%%%%%%%%%%%%%%%%%
%%%%%%%%%%%%%%%%%%%%%%%%%%%%%%%%%%%%%%%%%%%%%%%%%%%%%%%%%%%%%%%%%%%%%%%%%%%%%%%
\section{Lessons learned and what did not work}

Motivated by \citet{yang2026t1}, this section records problems we encountered and approaches that did not work, together with the lessons we drew from them.

\mypar{Differentiating the $k_3$ estimator gives forward-KL gradients.}
Delta-OPD regularizes the student toward its SFT reference with a differentiable reverse-KL term (\cref{eq:delta}), and our first implementation used the $k_3$ estimator~\citep{schulman2020kl} for that term, as is common in RL frameworks.
The $k_3$ estimate of the reverse-KL value is accurate, but its gradient with respect to the policy is the gradient of the forward KL, so training regularized the wrong divergence.
We corrected this with the $k_3^{+}$ construction, which keeps the $k_3$ value estimate while backpropagating the $k_2$ gradient, matching the reverse-KL gradient in expectation.
A KL estimator that enters the loss must therefore be validated at the gradient level, since value-level agreement is not sufficient.

\mypar{Optimizer numerics can mask regression.}
Early weak-to-strong Vanilla-OPD runs on an FSDP training backend showed no post-peak regression, while the same recipes on a Megatron backend regressed reliably.
The configurations differed in optimizer-state precision, BF16 under FSDP and FP32 under Megatron, and we suspect that the FP32 optimizer follows the small late-stage gradient signal more faithfully while BF16 rounding damps it.
We therefore ran the study on a single backend with fixed numerics, and caution that late-stage dynamics can be sensitive to such details.

\mypar{Sequence-level KL does not expose the regular regime.}
We first indexed trajectories by sequence-level reverse KL, following the coordinate of \citet{gao2023scaling}.
Under this coordinate the initial regime is not consistently regular across teacher--student pairs.
Token-mean KL is also the principled choice for OPD: the immediate-token objective in \cref{eq:opd} carries no return-to-go, so the update controls the conditional next-token divergence rather than the joint sequence divergence (\cref{sec:setup}).

\mypar{PPO and best-of-$n$ functional forms do not transfer.}
We initially fitted the regular regime with the forms of \citet{gao2023scaling}, the quadratic $ad-bd^{2}$ from best-of-$n$ and the subquadratic $ad-bd\log d$ from RL.
Across the grid the curvature terms are not statistically identifiable within the regular window, echoing the quadratic diagnostics in \cref{app:statistics}.
The linear law in $d$ describes the regular regime, and the heterogeneous tails outside it follow neither functional form.

\mypar{Full-vocabulary loss accelerates weak-to-strong regression.}
For Vanilla-OPD we compared minimizing the full-vocabulary reverse KL as a differentiable loss against the sampled-token policy-gradient objective of \cref{sec:preliminary}.
Under the full-vocabulary loss, post-peak regression in weak-to-strong pairs began earlier and was more severe.
We therefore use the sampled-token form throughout.

\mypar{No parametric law for the peak location.}
We attempted to model the location $d_{\mathrm{peak}}$ of the gold-score maximum as a function of student and teacher scale.
The observed locations are non-monotone across the grid (\cref{sec:scaling}), and no simple parametric family survived identifiability checks.
We therefore model the peak value $G_{\mathrm{peak}}$, the initial rate $m$, and the transfer extent $d_{\mathrm{transfer}}$, and report the maximum's location as an observed quantity.

\end{document}